\documentclass[10pt,twocolumn,letterpaper]{article}

\usepackage[pagenumbers]{cvpr} 

\usepackage{graphicx}
\usepackage{amsmath}
\usepackage{amssymb}
\usepackage{booktabs}
\usepackage{caption}
\usepackage{subcaption}

\usepackage[pagebackref,breaklinks,colorlinks]{hyperref}

\usepackage[capitalize]{cleveref}
\crefname{section}{Sec.}{Secs.}
\Crefname{section}{Section}{Sections}
\Crefname{x`table}{Table}{Tables}
\crefname{table}{Tab.}{Tabs.}

\newif\ifdraft
\draftfalse
\drafttrue

\usepackage[table]{xcolor}
\definecolor{orange}{rgb}{1,0.5,0}
\definecolor{violet}{RGB}{70,0,170}

\ifdraft
 \newcommand{\PF}[1]{{\color{red}{\bf PF: #1}}}
 
 \newcommand{\MS}[1]{{\color{blue}{\bf MS: #1}}}
 
 \newcommand{\AD}[1]{{\color{violet}{\bf AD: #1}}}
 
  \newcommand{\SH}[1]{{\color{cyan}{\bf SH: #1}}}
 
 \newcommand{\AR}[1]{{\color{teal}{\bf AR: #1}}}

\else
 \newcommand{\PF}[1]{}
 
 \newcommand{\MS}[1]{}
 
 \newcommand{\AD}[1]{}
 
  \newcommand{\SH}[1]{}
 
\newcommand{\AR}[1]{}

\newcommand{\AR}[1]{}

\fi

\newcommand{\comment}[1]{}
\newcommand{\parag}[1]{\vspace{-3mm}\paragraph{#1}}

\newcommand{\bod}[0]{{\cal B}}

\newcommand{\cG}[0]{{\mathcal G}}
\newcommand{\cD}[0]{{\mathcal D}}

\newcommand{\bx}[0]{{\bf x}}
\newcommand{\by}[0]{{\bf y}}
\newcommand{\bz}[0]{{\bf z}}

\newcommand{\ours}[0]{{\it Ours}}

\def\cvprPaperID{4380} 
\def\confName{CVPR}
\def\confYear{2022}

\begin{document}

\title{Adversarial Parametric Pose Prior}

\author{
Andrey Davydov\\
{\tt\small andrey.davydov@epfl.ch}
\and
Anastasia Remizova\\
{\tt\small anastasia.remizova@epfl.ch}
\and
Victor Constantin\\
{\tt\small victor.constantin@epfl.ch}
\and
Sina Honari\\
{\tt\small sina.honari@epfl.ch}
\and
Mathieu Salzmann\\
{\tt\small mathieu.salzmann@epfl.ch}\\\\
CVLab, EPFL
\and
Pascal Fua\\
{\tt\small pascal.fua@epfl.ch}
}

\maketitle


\begin{abstract}

The Skinned Multi-Person Linear (SMPL) model can represent a human body by mapping pose and shape parameters to body meshes. This has been shown to facilitate inferring 3D human pose and shape from images via different learning models. However, not all pose and shape parameter values yield physically-plausible or even realistic body meshes. In other words, SMPL is under-constrained and may thus lead to invalid results when used to reconstruct humans from images, either by directly optimizing its parameters, or by learning a mapping from the image to these parameters.

In this paper, we therefore learn a prior that restricts the SMPL parameters to values that produce realistic poses via adversarial training. We show that our learned prior covers the diversity of the real-data distribution, facilitates optimization for 3D reconstruction from 2D keypoints, and yields better pose estimates when used for regression from images. We found that the prior based on spherical distribution gets the best results. Furthermore, in all these tasks, it outperforms the state-of-the-art VAE-based approach to constraining the SMPL parameters.

\end{abstract}


\section{Introduction}
\label{sec:introduction}

The SMPL model~\cite{Loper15} is now widely used to parameterize body poses and shapes~\cite{Omran18,Li21e}. However, it offers no guarantee to produce realistic human bodies when random values are passed as its inputs. This complicates its usage within an optimization, regression, or generative frameworks, where it is desirable that any sample drawn be plausible. 

To mitigate this issue,  several approaches have been used. In ~\cite{Bogo16},  this is addressed by introducing  a Gaussian Mixture Model (GMM) learned on the SMPL pose. Unfortunately, due to its unbounded nature, it still allows poses far away from any training example and potentially unrealistic. In SMPL-X~\cite{Pavlakos19}, a Variational Autoencoder (VAE)  is used instead to learn a low-dimensional representation of the SMPL parameters. This choice was motivated by the ability of VAEs to model the distribution of valid data samples in the latent space as a multivariate Gaussian, which was shown to better approximate the data distribution than classical models, such as GMMs, while also facilitating sampling at test time. In both approaches, the learned prior is then used together with other losses in an optimization-based framework that aims at finding plausible human meshes.   Unfortunately, VAEs have drawbacks. First, their learned prior tends to be mean-centered and to discard part of the original data distribution that are far away from it. Furthermore, its Gaussian prior is unbounded, like that of GMMs. Hence, one can also sample latent values far away from any in the training set and produce unrealistic bodies. Adversarial training has been used to bound the parameter prediction of SMPL in regression-based frameworks~\cite{Kanazawa18a,Georgakis20}. However, this requires balancing the adversarial loss with other losses. More importantly, no explicit prior has been learned in such cases, as this training needs to be repeated for each new task.

In short, these approaches make it necessary to balance different losses and do not bound the inputs of the SMPL model. In contrast, we aim at learning a prior, that once learned can be used in an optimization or learning-based frameworks, without the requirement of enforcing constraints on it. In other words, the learned prior should be integrated as part of the SMPL model and the model can be optimized only on the target loss, where the learned prior is not added as an extra constraint. To this end, we learn an explicit prior, that constrains the input of the SMPL model to be realistic poses via adversarial training. This has to be done only once and independently of the target application so that no further adversarial training is needed. Hence, it does not require balancing multiple losses in the downstream tasks. 

Furthermore, one can use a bounded distribution, such as uniform or spherical ones, in the input space of the learned prior, which facilitates plausible sample generation and also its integration in regression frameworks, as by limiting the output of the preceding component that is passed as input to the learned prior, one is always guaranteed to have a plausible human representation. Once trained, our model can be used in many different settings without further retraining as shown in Fig.~\ref{fig:overview}. 
We introduce GAN-based pose prior learning technique that consistently outperforms the VAE-based state-of-the-art approach for both optimization- and regression-based approaches to human body pose recovery~\cite{Pavlakos19}. Also, we make a comparision between different choices of latent spaces, out of which the spherical one brings the most benefit.


\section{Related Work}
\label{sec:related}

\subsection{Body Representation}
There have been many attempts at modeling the human body. The earliest ones split the body into several simpler shapes and combine them into a unified model. The introduction of several datasets consisting of diverse body scans~\cite{Robinette02} has ushered the age of learnable body models. The SMPL body model~\cite{Loper15} constitutes one of the most successful and easy-to-use models. 
It uses a combination of PCA coefficients to model the shape and a regressor that poses the body from the joints angles. Several extensions have since then been proposed. SMPL-H~\cite{Romero17} includes a more detailed hand model, thus removing one of the limitations of the original model. 
More recently, SMPL-X~\cite{Pavlakos19}, adds facial expressions to the previous models. 
Instead of using a mesh representation, NASA~\cite{Deng20a} encodes the human body as a sign distance function. Here, we focus on the SMPL model as we are interested in modeling the human body itself, and favor a mesh representation which inherently provides correspondences across, e.g., video frames.

\subsection{SMPL Parameter Estimation}
Since the introduction of the SMPL body model, many approaches have aimed to estimate the SMPL parameters given either an image~\cite{Kanazawa18a, Pavlakos19a, Georgakis20, Jiang20c}, some labels, such as 2D or 3D pose~\cite{Pavlakos19, Bogo16, Arnab19}, or body silhouettes~\cite{Lassner17a}. Depending on whether they are optimization- or regression-based, they can be divided into three categories.

\paragraph*{Optimization Models.}
The first category consists of optimizing the SMPL parameters so as to minimize an objective function defined in terms of different pose or image descriptors. Such descriptors can be 2D and/or 3D joint locations~\cite{Bogo16, Arnab19, Pavlakos19}, silhouettes~\cite{Lassner17a}, or dense correspondences~\cite{Guler19}. SMPLify~\cite{Bogo16} constitutes one of the first such methods. It uses a GMM to model the pose space and optimizes the SMPL parameters so as to match 2D joint locations. The unboundedness of the GMM prior may result in the optimization producing unrealistic poses. Therefore, VPoser~\cite{Pavlakos19} proposed to replace the GMM with a VAE to model the pose space distribution. While a VAE can model more complicated distributions than a GMM, it remains unbounded. Furthermore, the mean-centered nature of VAEs makes it cover the original data distribution only partially, because it poorly represents data samples away from its distribution's means. As we will show later, our approach learns a better and smoother coverage of the data while addressing the unbounded nature of these approaches.

\paragraph*{Regression Models.}
The second category consists of directly regressing the SMPL parameters given an input image. Human Mesh Recovery (HMR)~\cite{Kanazawa18a} is one of the initial methods that applies such a technique using deep neural networks. Since then it has been used in several other works, such as~\cite{Georgakis20,Jiang20c,Pavlakos19a}. These methods minimize an adversarial prior together with other target losses.
Therefore, the resulting representation is only usable within the learned model, since no explicit prior is learned. By contrast, we learn an explicit bounded prior, which needs to be trained only once. Then, the weights of this learned prior can be frozen and it can be used in any regression approach by mapping a feature space to the learned prior latent space.

\paragraph*{Combined Models.}

The two previous categories are compatible with each other and can be used together. SPIN~\cite{Kolotouros19} mixes the two by fine-tuning the regression estimate with an optimization procedure. EFT~\cite{Joo20} takes the pretrained regression network of~\cite{Kolotouros19} and uses its weights as an implicit body prior. It fine-tunes the weights of the network for every sample in a weakly-annotated dataset to obtain the body parameters.  Although we demonstrate our method separately on optimization and regression-based tasks, it can be used in the combined approach, as these models merge the individual components from optimization and regression-based approaches.


\section{Method}
\label{sec:method}


\begin{figure*}
	\begin{center}
	 \includegraphics[width=\textwidth]{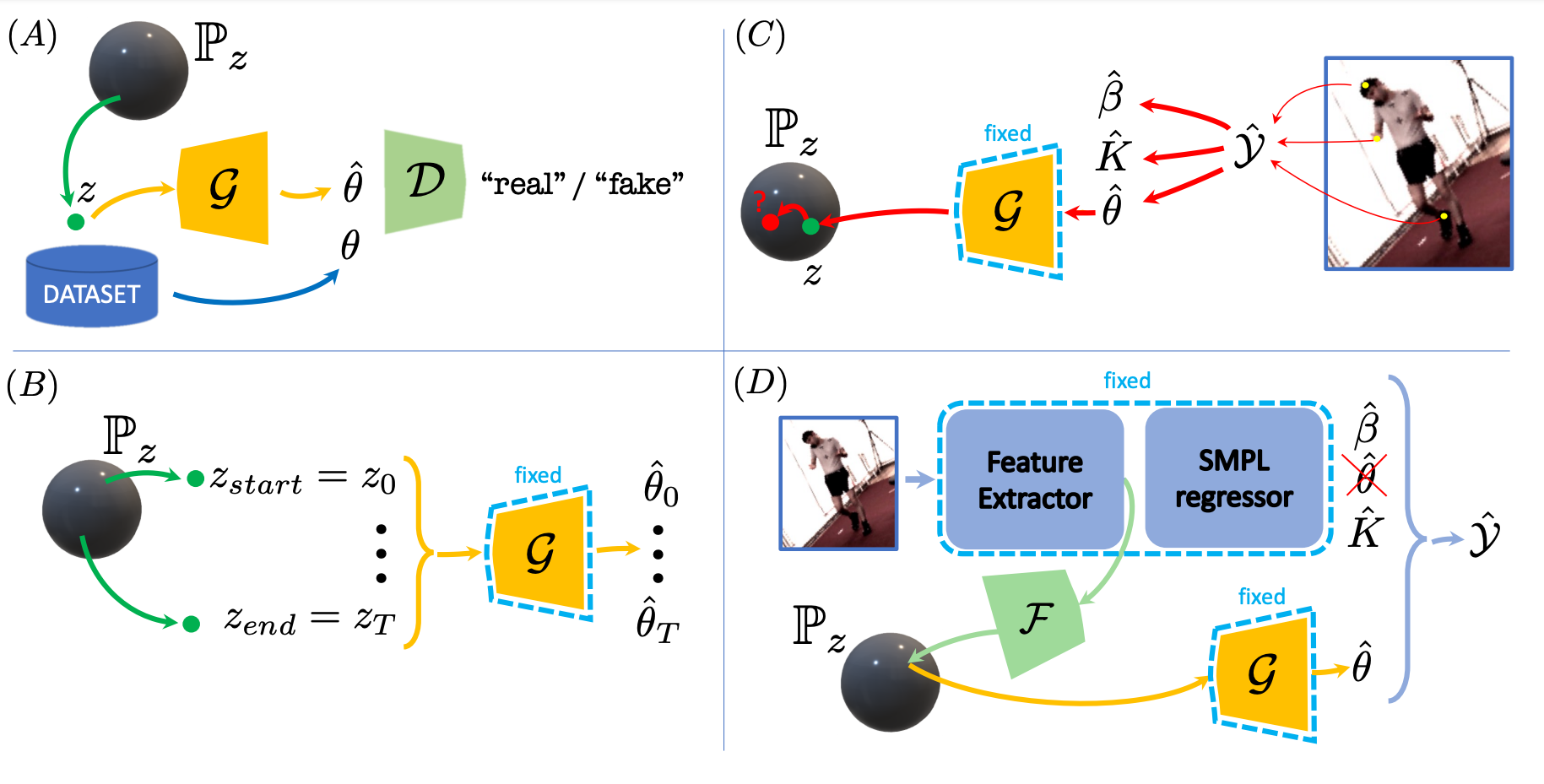}
	\vspace{-10mm}
	\end{center}
		 \caption{
			 Multiple uses of the proposed approach. 
			 $(A)$ The adversarial game between generator $\mathcal{G}$ and discriminator $\mathcal{D}$ guarantees that the former provides realistic body poses. 
			 $(B)$ Given ``start" and ``end" latent vectors, one can render the whole sequences of plausible and smooth body interpolations. 
			 $(C)$ To optimize for corresponding poses given target keypoints $\mathcal{Y}$, akin to VPoser~\cite{Pavlakos19}, the pretrained generator $\mathcal{G}$ can be used as an implicit pose prior.  
			 $(D)$  The pretrained $\mathcal{G}$ can be dropped in as a pose prior in a pretrained off-the-shelf human mesh regressor. It improves the prediction quality of the regressor.
			 }
	\label{fig:overview}
\end{figure*}

To constrain the SMPL poses we rely on a GAN approach~\cite{Goodfellow14}. It involves two competing networks, a {\it generator} $\cG$ and a {\it discriminator} $\cD$. The generator samples vectors $\bz$, known as {\it latent vectors},  from a set $\mathbb{P}_z \subseteq \mathbb{R}^d$ and generates a SMPL pose vector $\hat{\Theta} = \mathcal{G}(\bz)$, which can be passed to the SMPL decoder $\phi$ to generate a body mesh  $\bod = \phi((\bz),\beta)$, where $\beta$ denotes the SMPL body shape parameters. The task of the discriminator is to distinguish poses generated in this manner from those of a large dataset of poses known to be realistic. By contrast, the generator is trained to produce poses that fool the discriminator. This process is shown in Fig.~\ref{fig:overview} $(A)$.

\parag{Constraining shape and pose}

Training our models, we leave the SMPL shape parameters $\beta$ untouched to be able to compare with other models, i.e. VPoser~\cite{Pavlakos19}, as they only learn a prior for pose. 
Moreover, the  shape part of the model is already data-driven (with PCA). However, the PCA weights for the shape are also unbounded by the model and eliminating this problem is also worth further research. We trained a model in such combined fashion, more information on this can be found in the supplementary material.

\subsection{Distribution over Latent Vectors}

GAN-based approaches~\cite{Ulyanov18, Niemeyer21, Karras19} have used several types of distributions from which to draw their latent vectors, including Gaussian, Uniform, and Spherical distributions. To test all three, we learn three different sets of latent vectors:
\begin{itemize}
    \item GAN-N:  $\bz_N \sim \mathbb{P}_z = \mathcal{N}(0, \mathcal{I})$ (\textit{Normal})
    \item GAN-U:  $\bz_U \sim \mathbb{P}_z = \mathcal{U}(-1,1)$ (\textit{Uniform})
    \item GAN-S:  $\bz_S \sim \mathbb{P}_z = \mathcal{S} \subset \mathbb{R}^d$  (\textit{Spherical})
\end{itemize}
where the spherical vectors are sampled by drawing vectors $\bz_N$ from a normal distribution and computing $\bz_S := \frac{\bz_N}{\|\bz_N\|_2}$.

The unbounded nature of the Gaussian distribution $\mathcal{N}(0, \mathcal{I})$ prevents sampling from rare modes and may make the resulting prior suffer from the same drawbacks as GMMs and VAEs when used in regression tasks. While the Uniform distribution does not have such a limitation, it imposes artificial bounds $[-1,1]^d$ that do not have a clear meaning in the output pose space. Intuitively, because one can smoothly move from one pose to another, we would rather expect a latent pose space to be continuous, without strict boundaries as the uniform space. The desirable properties of the latent space, such as continuity and boundedness are all inherent to the Spherical distribution. Our experiments show that, in practice, it does indeed tend to perform better than the others.

\subsection{Training}

We define the generator $\mathcal{G}$ in our GAN architecture to have the same structure as the decoder of the VAE in VPoser~\cite{Pavlakos19}. As for our discriminator $\mathcal{D}$, we base our structure on that of the HMR approach~\cite{Kanazawa18a}, using $K+1$ discriminators, one for each joint angle and one for the whole set of pose parameters.

As can be seen in Fig.~\ref{fig:overview} $(A)$, we draw samples from the latent space $\mathbb{P}_\bz$ and train the generator to map them to the SMPL pose space. The discriminators are trained to distinguish the SMPL pose vectors $\Theta$, obtained by a real pose dataset, from the ones produced by the generator $\hat{\Theta}$. The training loss function, aiming to balance the two opposing goals of the generator and discriminator, can thus be expressed as
\begin{align}
    \label{eq:gan}
\min_{\mathcal{G}}\max_{\mathcal{D}}\mathcal{L}(\mathcal{G}, \mathcal{D}) &= \mathbb{E}_{\Theta}[\log\mathcal{D}(\Theta)] +\\ \nonumber
&+\mathbb{E}_{z\sim\mathbb{P}_z}[\log(1-\mathcal{D}(\mathcal{G}(z))]\;.
\end{align}
In our experiments, we update the discriminator weights $10$ times for every update of the generator. As we found, it leads to the best performance.

We train this model using the common splits of the AMASS dataset~\cite{Mahmood19}, following the procedure for VPoser in SMPL-X~\cite{Pavlakos19}. As AMASS provides SMPL-H parameters~\cite{Romero17}, the body pose is a bit different from the original SMPL~\cite{Loper15, Bogo16} one; it only contains $K=21$ joint angles, with 2 angles from SMPL having been moved to SMPL-H ``hands" articulations.  


\subsection{Using the Generator as a Universal Prior}
\label{sec:applications}

Being trained once, our generative model can be used in many applications. We introduce some of them here and present the results in the following section.

\parag{Interpolation in Latent Space.} \label{par:interpolation}

One way to gauge the quality of a latent representation is to check how smooth the interpolation from one latent vector to another is. Ideally, the transition should vary equally in each step from the source to the target samples, rather than most of the transformation occurring in only a few steps. To check this, we randomly select $N$ samples $\{\bod{}^t_1, \ldots, \bod{}^t_N\}$ from the test set and optimize (similar to the paragraph below) for every body $\bod{}^t_i$ the corresponding latent vector $\bz^r_i$ that yields the closest mesh $\bod{}^r_i = \phi(G(\bz^r_i), \beta)$. For each pair of such latent vectors, we construct an interpolation sequence $\{\bz^r_0,  \bz^r_1, \ldots , \bz^r_T\}$ and $\{\bod_0,  \bod_1, \ldots , \bod_T\}$ using either linear interpolation
\begin{align}
\bz^r_t &= \big(1 - \frac{t}{T} \big) \bz^r_0 + \frac{t}{T}\bz^r_T \label{eq:lerp}
\end{align}
for GAN-N and GAN-U, or spherical interpolation
\begin{align}
    \bz^r_t &= \frac{\sin{\big(1 - \frac{t}{T}\big)\theta}}{\sin{\theta}} \bz^r_0 + \frac{\sin{\frac{t}{T}\theta}}{\sin{\theta}} \bz^r_T
    \label{eq:slerp}
\end{align}
for GAN-S, with $\bod{}^r_t = \phi(G(\bz^r_t), \beta)$, and $\theta$ representing the angular distance between two points $\bz^r_0$ and $\bz^r_T$ on a sphere. We discuss spherical interpolation (SLERP) in more detail, including the proof of Eq.~\ref{eq:slerp} for high dimensions, in the supplementary material.
Ideally, the samples $\bod_{t-1}$ and $\bod_{t+1}$ should be roughly equidistant from $\bod_{t}$, indicating smooth transitions. The per-vertex mesh distance is computed using Eq.~\ref{eq:m2m_dist}. The sampling process is depicted in Fig.~\ref{fig:overview} $(B)$.

\parag{Optimization from Keypoints.}

Given the 2D joint targets $\mathcal{Y}$ obtained from a monocular observation 
and assuming neutral SMPL shape parameters $\beta=0$,
our goal is to find the SMPL pose parameters $\hat{\Theta}$ that produce the target $\mathcal{Y}$ using the SMPL model $\phi$,
which translates from SMPL space $(\hat{\Theta}, \beta)$ to the space of body meshes $\bod$. Fig.~\ref{fig:overview} $(C)$ describes the idea.
The recovered mesh can be projected to 2D joints using camera parameters,
i.e., $\Pi(\phi(\hat{\Theta}, \beta)) = \mathcal{Y}$, where $\Pi$ is the camera projection function.
To find the optimal SMPL parameters, one can minimize $L(\Pi(\phi(\hat{\Theta}, \beta)), \mathcal{Y})$,
where $L$ is a loss function such as the $L2$ distance between the 2D mesh joints and the corresponding target joints.

To better constrain the pose output by SMPL, we make use of our pose prior. That is, instead of directly optimizing $\hat{\Theta}$, we optimize a vector $\bz$ in the GAN's latent space and obtain the corresponding $\hat{\Theta}$ by feeding $\bz$ to the generator $\mathcal{G}$. Altogether, we therefore solve the optimization problem
\begin{align}
\min_{\bz}\|\Pi(\phi(\mathcal{G}(\bz), \beta) - \mathcal{Y}\|^2_2\;.
\end{align}

\parag{Image-to-Mesh Regression.}

 Our GAN models can also be used as drop-in priors to improve existing pretrained image-to-mesh algorithms~\cite{Kanazawa18a, Kolotouros19, Joo20}. To demonstrate this, we start from the model of ~\cite{Joo20}, whose architecture is a \textit{Resnet50} model based on the one of~\cite{Kanazawa18a}. It is pretrained on pseudo ground-truth COCO~\cite{Lin14a} dataset obtained by ~\cite{Joo20}. We then inject our model into it as shown in Fig.~\ref{fig:overview} $(D)$. More specifically, we introduce an additional MLP $\mathcal{F}$ that maps intermediate features of \textit{Resnet50} to a latent vector $\bz$ of the pre-trained SMPL prior, which then can be mapped by $\mathcal{G}(\bz)$ into the pose vector $\Theta$ of SMPL. One can then decode pose parameters $\Theta$ into a human mesh $\bod$ using the SMPL model $\phi$. In turn, $\mathcal{F}$ can be used in conjunction with the pre-trained SMPL prior $\mathcal{G}$ and the SMPL decoder $\phi$ to reconstruct a complete body mesh, which can then be compared to the ground-truth targets. We used this process to train only the $\mathcal{F}$ in an end-to-end setup and obtain the corresponding body mesh $\bod$.
 

\section{Experiments}
\label{sec:experiments}

We now compare the three versions of our approach in sampling the latent vectors, GAN-N, GAN-U, and GAN-S, with the VPoser VAE-based approach of SMPL-X~\cite{Pavlakos19}. 


\subsection{Dataset Coverage}
\label{subsec:coverage}


\begin{table}[t]
	
	\caption{Statistics for Recall experiment on \textit{Train}/\textit{Test} splits of AMASS dataset ~\cite{Mahmood19}. The values are the mean, variance and medians of distances between real samples and closest neighbors among generated samples for every model. GAN models demonstrate indistinguishable behaviour, while VPoser~\cite{Pavlakos19} provides consistently larger discrepancy with real set.}
	\vspace{-4mm}
	\begin{small}
	\begin{center}
			\begin{tabular}{c|cc|cc}
				& \multicolumn{2}{c|}{\textit{Train} set} &  \multicolumn{2}{c}{\textit{Test} set}\\
				& $\mu \pm \sigma$ ($\downarrow$) & median ($\downarrow$) & $\mu \pm \sigma$ & median\\
          \hline
				GAN-S (\ours{}) &4.0$\pm$1.9           & 5.5          & 6.3$\pm$2.6 & 5.5 \\
				GAN-U (\ours{}) & \textbf{3.9}$\pm$\textbf{1.9} & 5.4          & \textbf{6.2}$\pm$\textbf{2.5} & \textbf{5.4} \\
				GAN-N (\ours{}) & 4.0$\pm$1.9     	   & \textbf{3.6} & \textbf{6.2}$\pm$\textbf{2.5} & 5.6 \\
				VPoser~\cite{Pavlakos19} 			& 5.2$\pm$3.2          & 4.3          & 6.3$\pm$4.0 & 7.3 

		\end{tabular}
	\end{center}
	\end{small}
	\label{table:recall}
\end{table}



\begin{figure}[htbp]
    \begin{subfigure}[b]{0.5\linewidth}
        \centering
        \includegraphics[width=4cm]{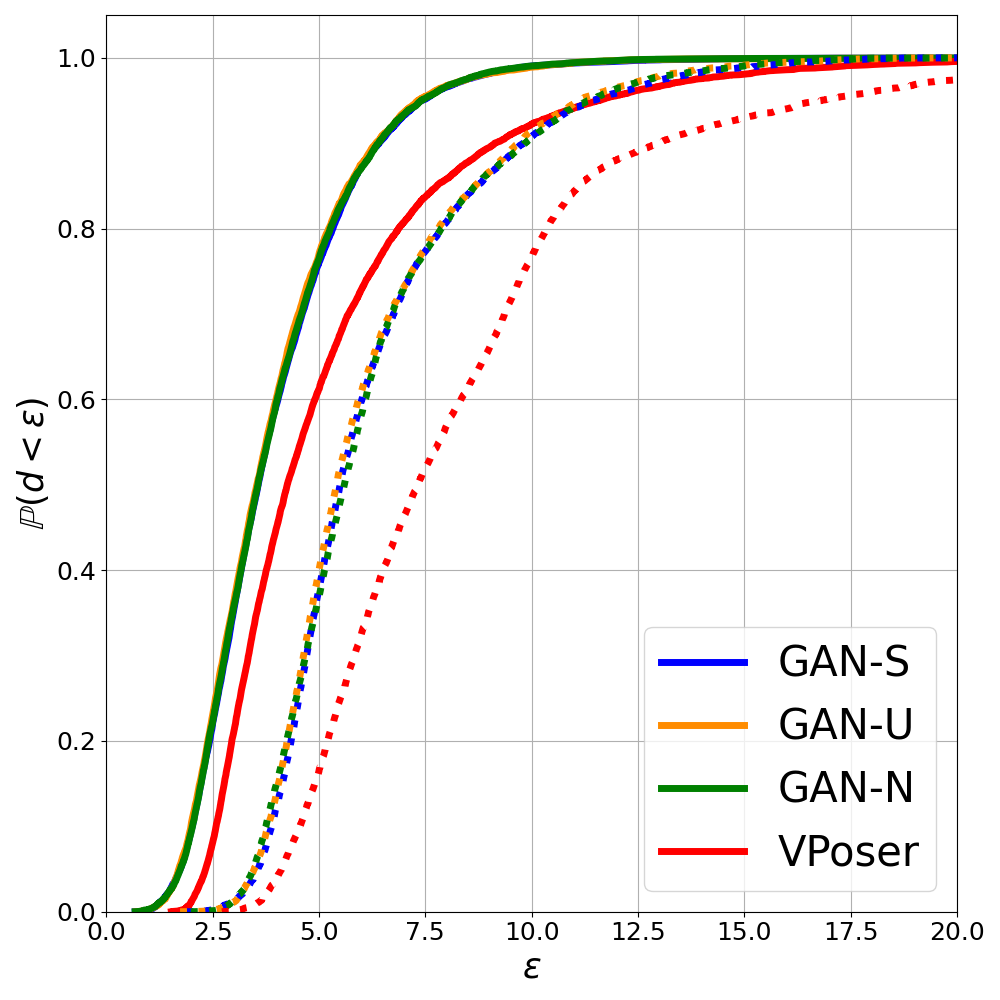}
        \caption{\centering Recall. ``Do \textit{real} samples live in latent spaces?"}\label{fig:recall}
       \end{subfigure}%
    \begin{subfigure}[b]{0.5\linewidth}
     \centering
     \includegraphics[width=4cm]{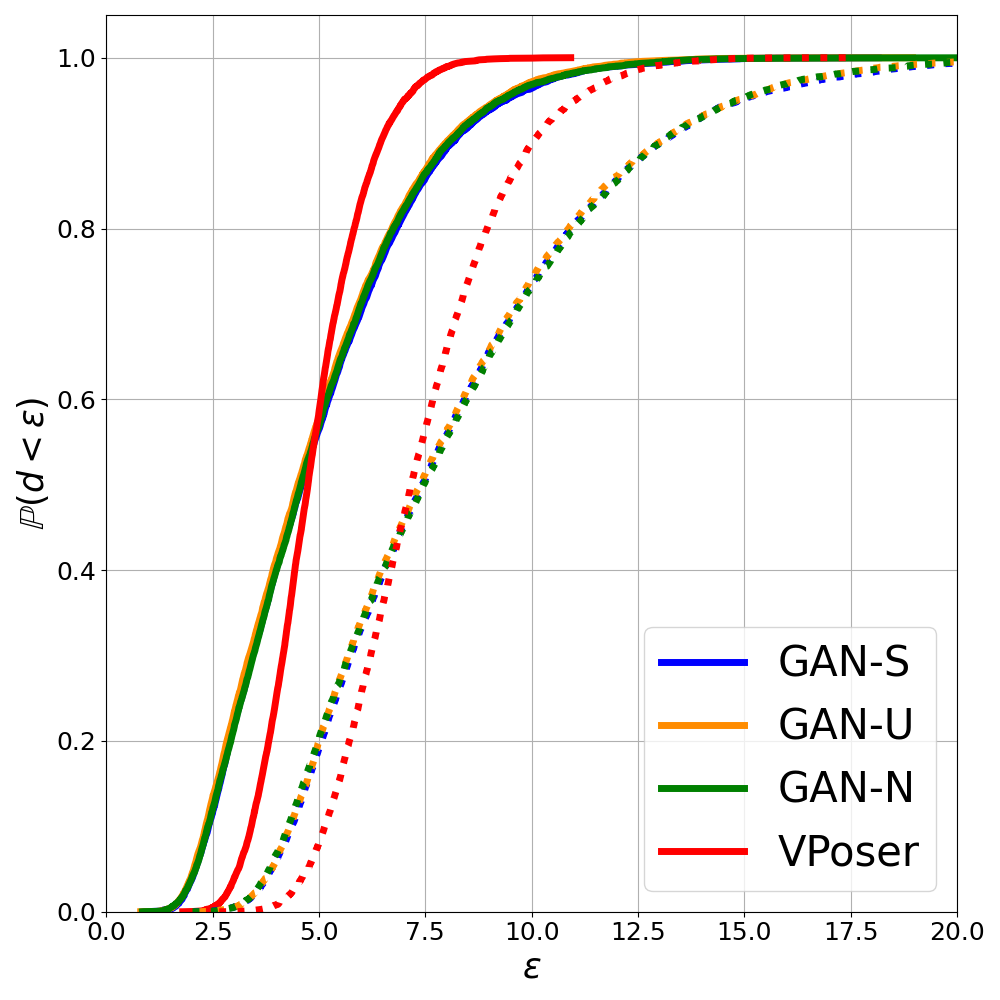}
     \caption{\centering Precision. ``How close are \textit{fake} samples to \textit{real}?"}\label{fig:precision}
    \end{subfigure}%
    \caption{Empirical estimation of data coverage of generative models for both \textit{Recall} (a) and \textit{Precision} (b). Experiments with data from the \textit{Train} set are drawn with solid lines, and from the \textit{Test} set with dashed lines. Higher means better in all charts.}
   \label{fig:precision_recall}
\end{figure}


\begin{figure*}%
    \centering
    \subfloat[\centering t-SNE on GAN-S samples]{{\includegraphics[width=0.2\textwidth]{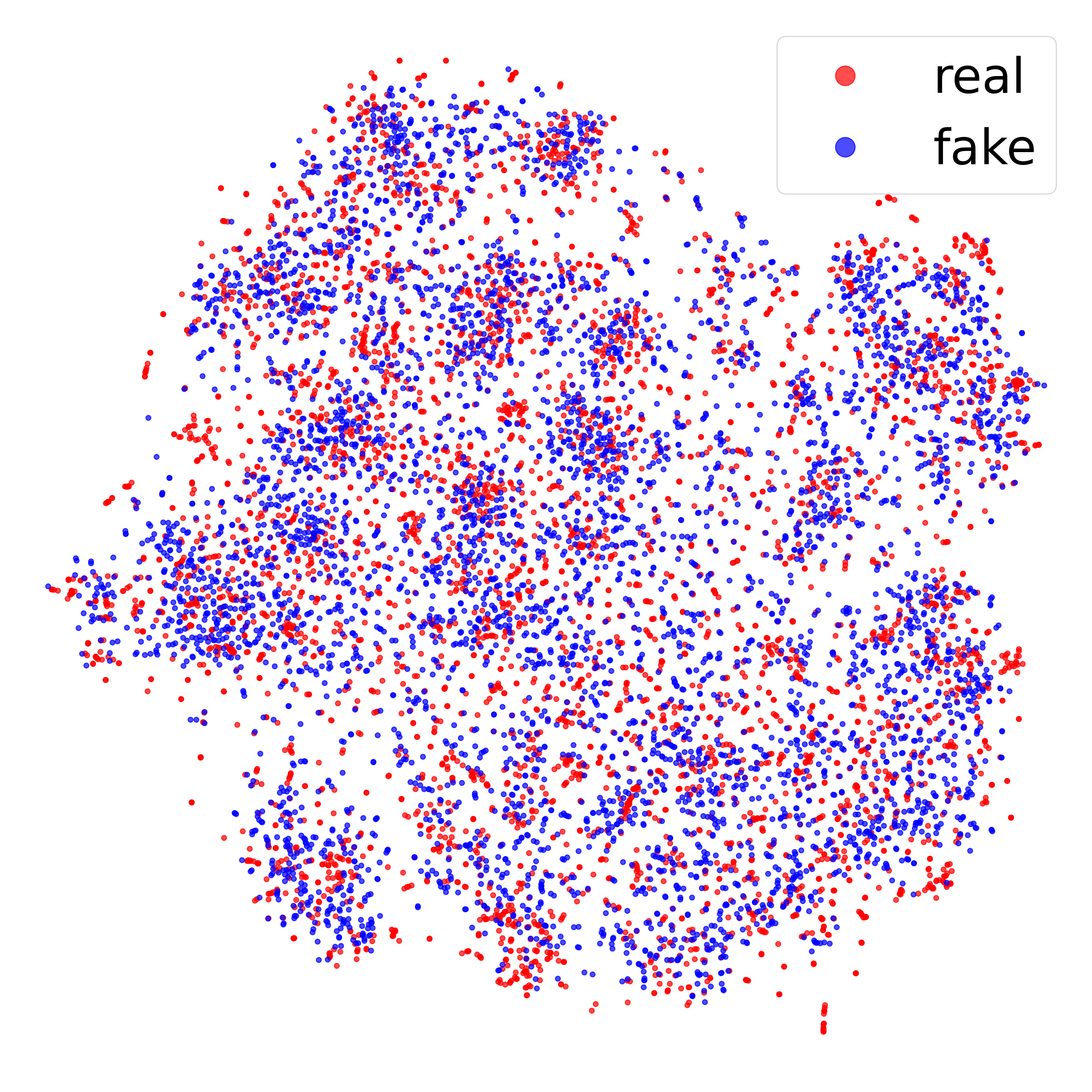} }}%
    \hspace{0.01cm}
    \subfloat[\centering t-SNE on GAN-U samples]{{\includegraphics[width=0.2\textwidth]{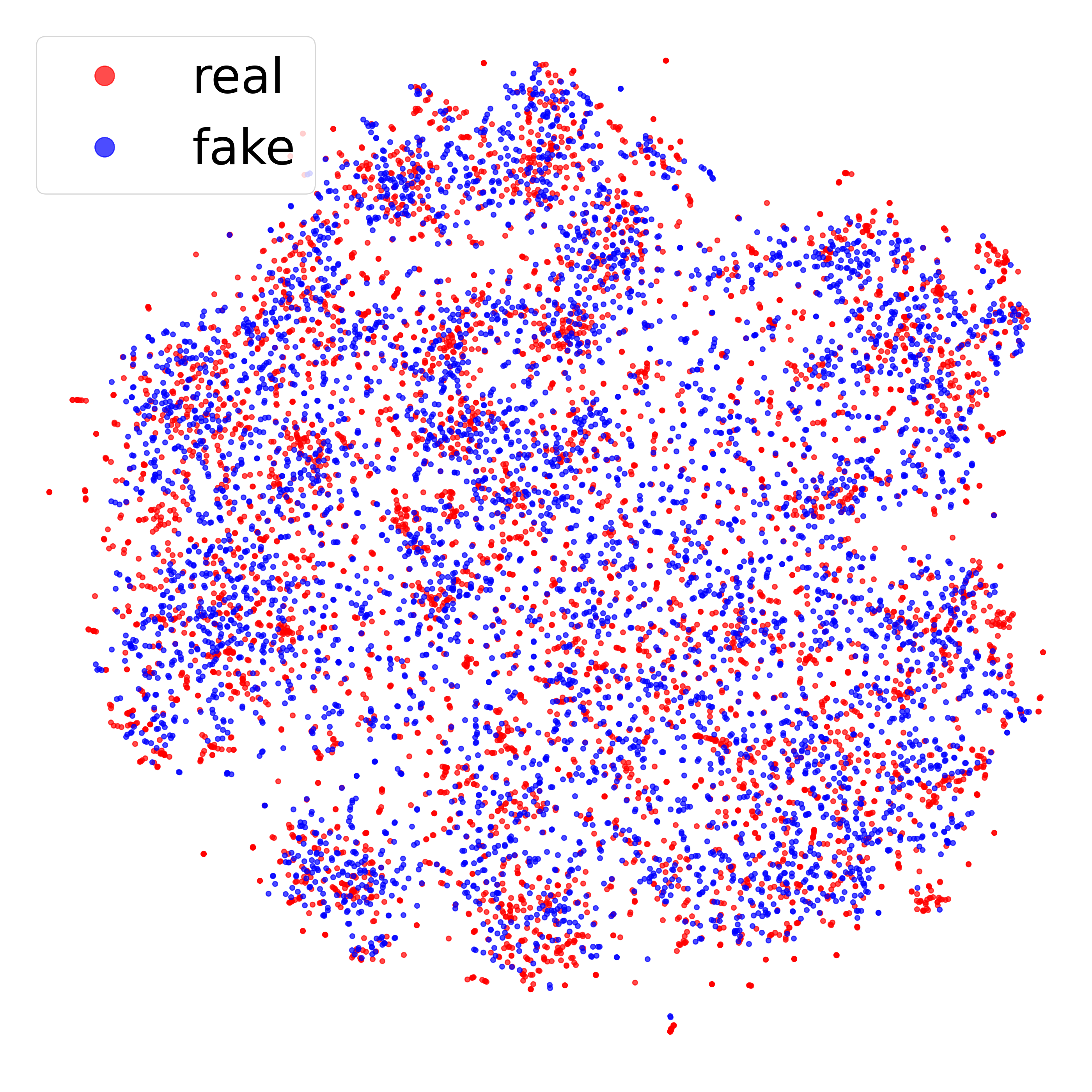} }}%
    \hspace{0.01cm}
    \subfloat[\centering t-SNE on GAN-N samples]{{\includegraphics[width=0.2\textwidth]{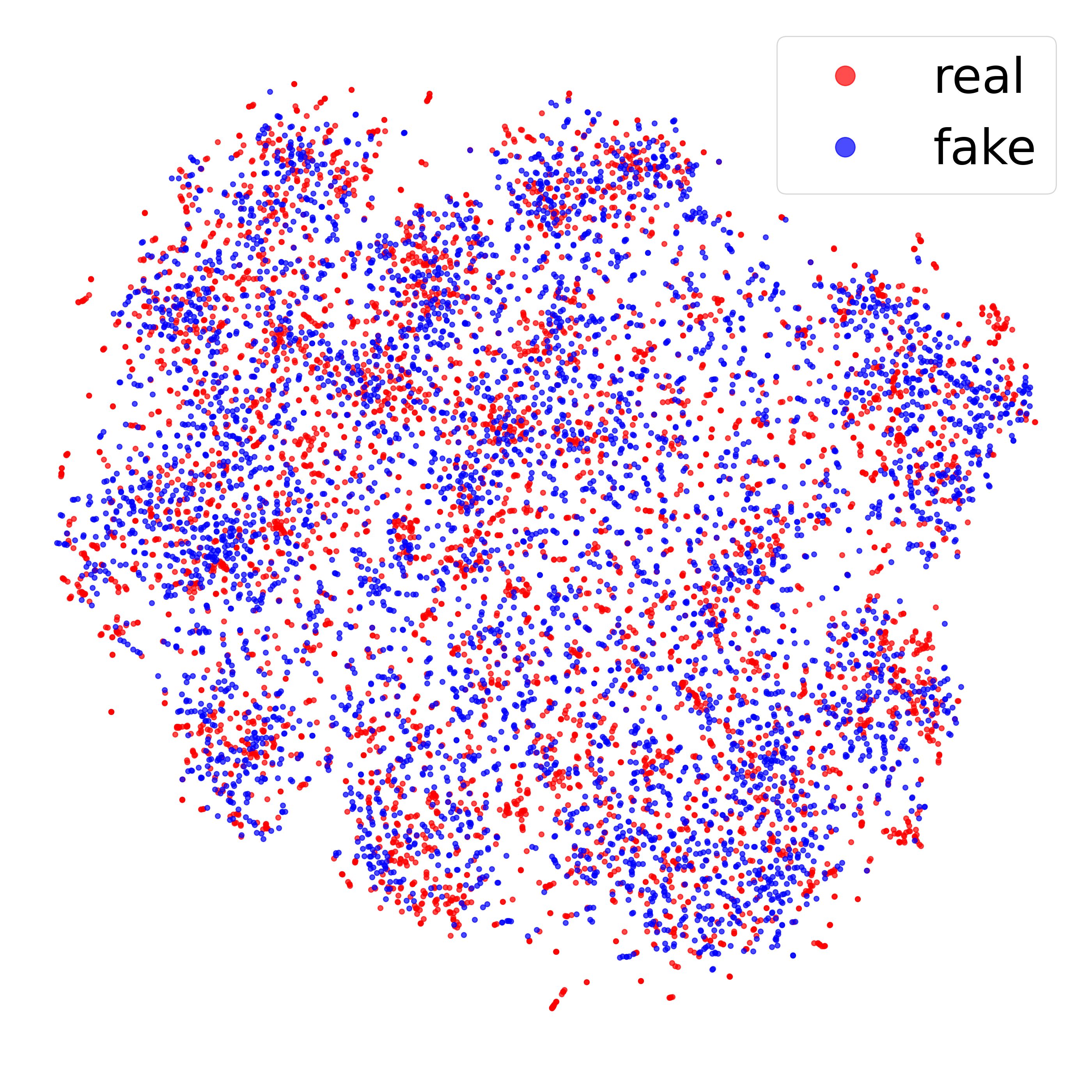} }}%
    \subfloat[\centering t-SNE on VPoser~\cite{Pavlakos19} samples]{{\includegraphics[width=0.2\textwidth]{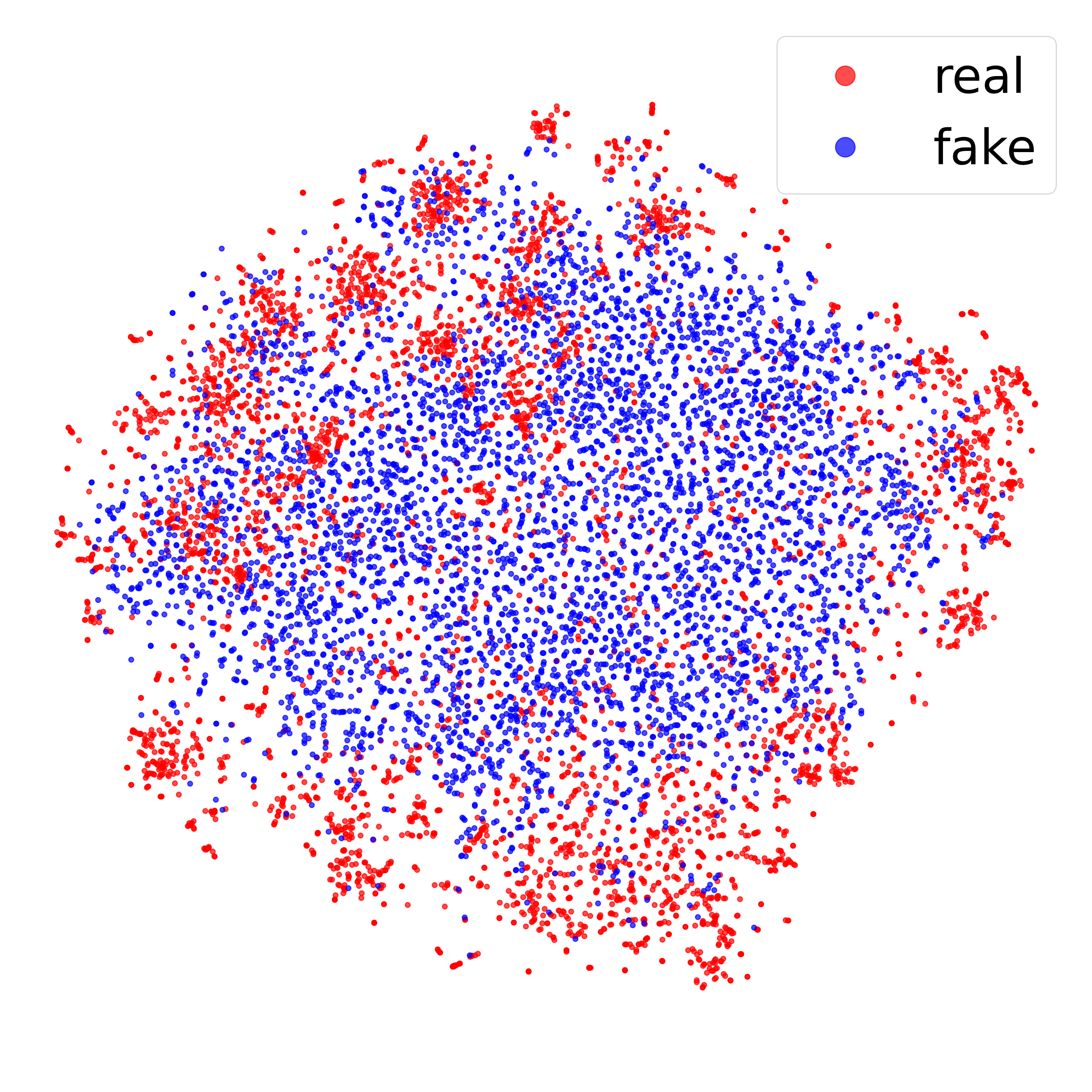} }}%
    \hspace{0.01cm}
    \caption{t-SNE projections of ``real" samples from the training set and of ``fakes" generated by GAN-S (a), GAN-U (b), GAN-N (c) and VPoser~\cite{Pavlakos19} (d) models. }%
    \label{fig:tsne}%
\end{figure*}

\begin{figure*}[t]
    \begin{center}
    \centering
    \fboxsep=0mm
    \fboxrule=1pt
    \vspace{.4em}
    \rotatebox{90}{\begin{minipage}[b]{40pt}{\begin{center} \footnotesize\text{GAN-S}\\\text{sample}\end{center}}\end{minipage}}\,{\includegraphics[width=0.8\textwidth]
    {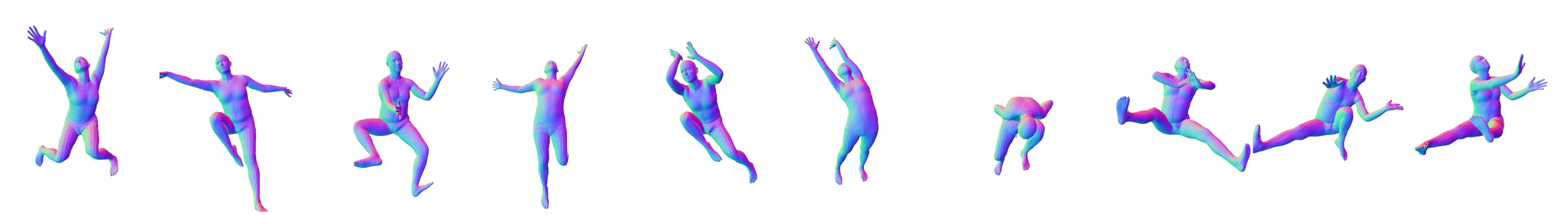}}\\

    \fboxsep=0mm
    \rotatebox{90}{\begin{minipage}[b]{40pt}{\begin{center} \footnotesize\text{GAN-S}\\\text{1NN}\end{center}}\end{minipage}}\,{\includegraphics[width=0.8\textwidth]
    {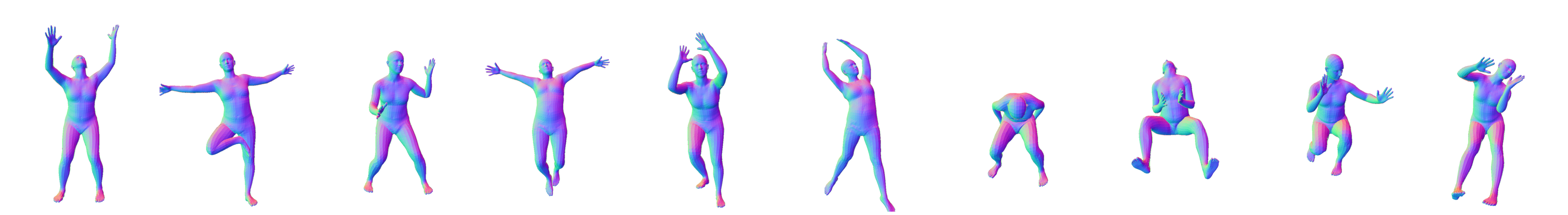}}\\

    \noindent\rule{\linewidth}{0.4pt}
    \fboxsep=0mm
    \rotatebox{90}{\begin{minipage}[b]{40pt}{\begin{center} \footnotesize\text{GAN-U}\\\text{sample}\end{center}}\end{minipage}}\,{\includegraphics[width=0.8\textwidth]
    {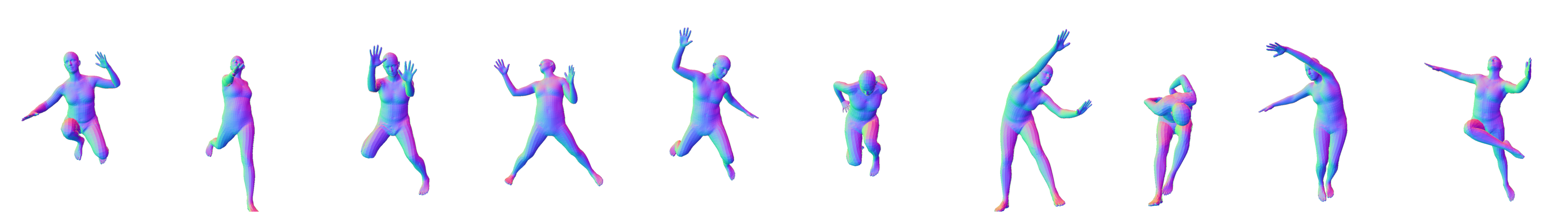}}\\

    \fboxsep=0mm
    \rotatebox{90}{\begin{minipage}[b]{40pt}{\begin{center} \footnotesize\text{GAN-U}\\\text{1NN}\end{center}}\end{minipage}}\,{\includegraphics[width=0.8\textwidth]
    {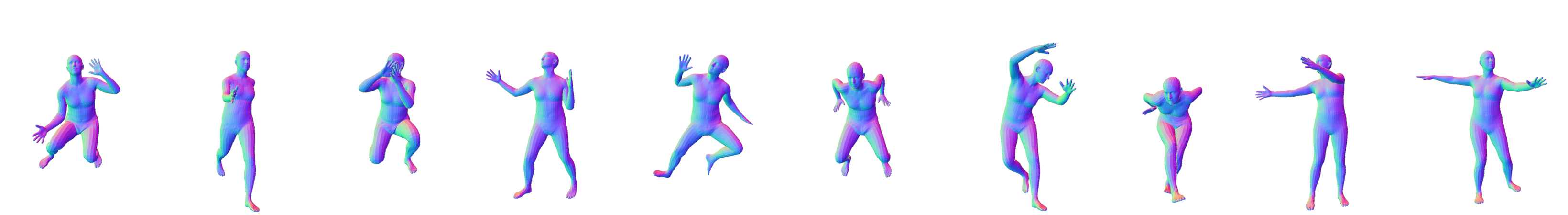}}\\

    \noindent\rule{\linewidth}{0.4pt}
    \fboxsep=0mm
    \rotatebox{90}{\begin{minipage}[b]{40pt}{\begin{center} \footnotesize\text{GAN-N}\\\text{sample}\end{center}}\end{minipage}}\,{\includegraphics[width=0.8\textwidth]
    {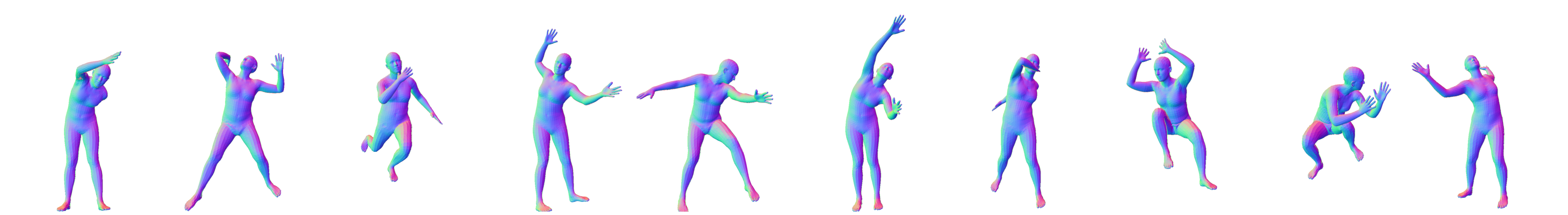}}\\

    \fboxsep=0mm
    \rotatebox{90}{\begin{minipage}[b]{40pt}{\begin{center} \footnotesize\text{GAN-N}\\\text{1NN}\end{center}}\end{minipage}}\,{\includegraphics[width=0.8\textwidth]
    {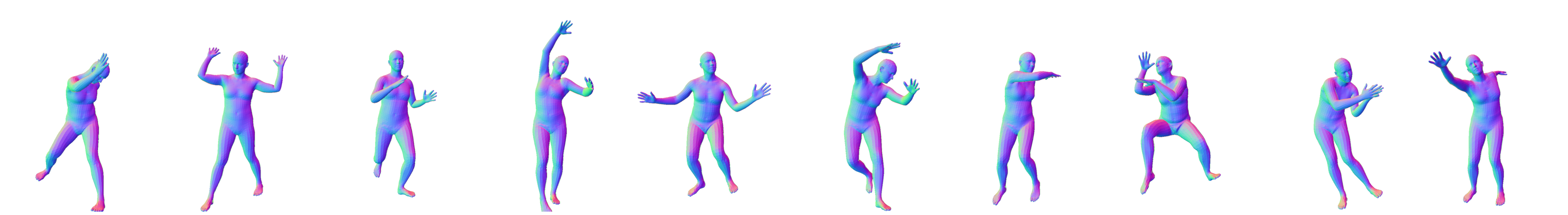}}\\

    \end{center}
    \vskip-6pt
        \caption{``Worst" samples according to the precision metric In Fig.~\ref{fig:precision}. For each GAN model we show 10 samples with the largest distance to the first nearest neighbor (1NN) in the training set, ordered from the worst sample on the right. Generated samples themselves are absolutely plausible human bodies, despite being away from training samples. Note that in GAN-S and GAN-U these samples are further away from 1NN compared to GAN-N. 
    }
    \label{fig:worst}
\end{figure*}

An ideal latent representation should cover the whole space of realistic human poses and nothing else. In other words, it should have good {\it Recall} and {\it Precision}. By recall, we mean that all samples in the training set should be well approximated by poses our model generates. By precision, we mean that these generated poses should never deviate too far from the training set. While recall indicates how well the generated samples cover the dataset distribution, precision indicates how realistic the generated samples are. We define these metrics as follows.

\parag{Recall.}

To evaluate recall, we use our pose generator to produce SMPL poses and take the shape parameters to be a zero vector, which yields a neutral body shape. Hence, for all models we produce a body mesh $\bod =\phi(\mathcal{G}(\bz), \beta)$ given a sampled latent vector $\bz$ and a fixed $\beta$. We first generate \textit{6M} samples from the pose generator of each model. Then, given a ground-truth body $\bod^t$ from either the training or test set, we select the generated body $\bod^r =\phi(\mathcal{G}(\bz^r), \beta)$  with minimum vertex-to-vertex distance

\begin{align}
    d(\bod{}^r, \bod{}^t) &= \frac{1}{N_{verts}}\sum_{v=1}^{N_{verts}} \| \bod{}^r_{v} - \bod{}^t_{v} \|_2 \; ,
    \label{eq:m2m_dist}
\end{align}

where $v$ sums over the vertices.

We then repeat this operation for \textit{10k} bodies randomly sampled from either the training or test set. We report the mean, variance and median of the resulting distances in Table \ref{table:recall}. In Fig.~\ref{fig:recall}, we plot the cumulative distribution $\mathbb{P}(d < \epsilon)$ given the values $d(\bod{}^r, \bod{}^t)$ for each training sample. Note that all versions of our approach deliver consistently higher values than the VPoser~\cite{Pavlakos19}, indicating that our models better cover the entire distribution.

In Fig.~\ref{fig:tsne}, we show the t-SNE projection~\cite{Maaten08} of the resulting  SMPL $\Theta^r$ pose vectors superposed on the $\Theta^t$ vectors that were used to generate the training examples. All GANs cover the space spanned by the training examples more completely than VPoser, which is consistent with the previous result. In other words, our learned prior can represent more diverse poses than the other ones.

\parag{Precision.}

Our approach to computing precision mirrors the one we used for recall. We randomly generate \textit{10k} latent points $\bz^r$ from every model, and for each sample $\bod{}^r = \phi(G(\bz), \beta)$, with a fixed $\beta$, look in the training or test datasets for the nearest neighbor in terms of the distance given in Eq~\ref{eq:m2m_dist}. If the latent representation only produces poses similar to those seen in training, this distance should be consistently small.

As shown in Fig.~\ref{fig:precision}, GAN models tend to produce meshes that are further away from the training distribution than the VAE model. This could be interpreted as a failure to produce realistic poses. However, these unseen samples correspond to plausible bodies. They are nothing but the result of semantic interpolation that GANs implicitly learn from the data. In Fig.~\ref{fig:worst}, we show the worst 10 samples based on the distance metric of Eq.~\ref{eq:m2m_dist} and their closest nearest neighbors from the training set. Note that all of these samples look realistic even though they are far from the closest neighbor in the dataset. This indicates that our generators are able to produce novel samples that were not observed in the training set, however, this is more observed in GAN-S and GAN-U compared to GAN-N, as GAN-N generates samples closer to its mean, hence deviating less to more diverse poses.


\subsection{Interpolation in Latent Space}

To evaluate our model on the first application described in Section~\ref{sec:applications} and in Fig.~\ref{fig:overview} $(B)$, we randomly select $N=128$ samples $\{\bod{}^t_1, \ldots, \bod{}^t_N\}$ from the test set, and, for each pair, we construct the corresponding interpolation sequence $\{\bz^r_0,  \bz^r_1, \ldots , \bz^r_T\}$ and $\{\bod_0,  \bod_1, \ldots , \bod_T\}$. We use the mean per-vertex position error (Eq.~\ref{eq:m2m_dist}) between body meshes $\bod_i$ and $\bod_j$ and compute pairwise distances between their body meshes by $\Delta(\bod_i, \bod_j)$, which we represent by $\Delta_{ij}$.

The minimal transformation $\Delta_{ij}$ between every consecutive bodies is equal to $\Delta_{0T} = \frac{\Delta(\bod^{i}_0, \bod^{i}_T)} {T}$. For different initial pairs, this value can be drastically different, as the corresponding bodies might be very close to or very far from each other. Hence, we normalize the transformation $\Delta_{ij}$ of every sequence by the expected average transformation $\Delta_{0T}$, yielding $\bar{\Delta}_{ij}$, which should be $1$ in the minimal case. Note, however, that such an ideal case can typically only be achieved by going through physically-impossible poses, for instance by shrinking the arms to go from a body with arms up to one with arms down.
Hence, actual transformations will typically obtain values higher than $1$, but a good latent space should nonetheless yield values as constant as possible throughout the entire interpolation steps, indicating a smooth gradual transition. 


\begin{figure}
	\begin{center}
	 \includegraphics[width=.47\textwidth]{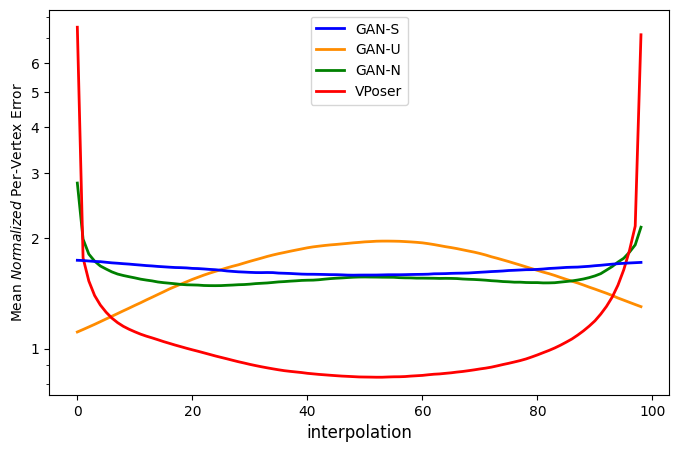}
   \vspace{-20pt}
	\end{center}
		 \caption{Average normalized mesh transformation of consecutive generated samples in interpolations between source and target examples. $x$ axis depicts the number of iterations and the $y$ axis shows the normalized mesh deviation in log-scale. The ideal transformation should be represented as a flat line. 
		 VPoser applies most of transformations at the beginning or at the end of the interpolations. GAN-N reduced this affect, while GAN-S obtains the smoothest transitions.}
	\label{fig:interpolation}
\end{figure}


\begin{table}[t]
	
	\caption{Comparison of interpolation smoothness between different models. $max$ to $min$ ratios of consecutive deformations for every pair of interpolation is measured according to Eq.~\ref{eq:interp_norm}.  Mean, median, and standard deviation of $R_{ij}$ over a set of interpolations is then reported for each model. }
	\vspace{-4mm}
	\begin{small}
	\begin{center}
			\begin{tabular}{c|cc}
				  & $\mu \pm \sigma$ ($\downarrow$) & median ($\downarrow$) \\
          \hline
				GAN-S (\ours{}) & \textbf{4.3}$\pm$\textbf{3.0}         & \textbf{3.5} \\
				GAN-U (\ours{}) & 7.1$\pm$7.7 		                    & 5.1 \\
				GAN-N (\ours{}) & 5.7$\times10^2\pm$5.3$\times10^2$     & 8.6 \\
				VPoser~\cite{Pavlakos19} 			& 1.8$\times10^5\pm$6.9$\times10^5$     & 9.1

		\end{tabular}
	\end{center}
	\end{small}
	\vspace{-6mm}
	\label{table:interp}
\end{table}


\begin{figure*}[t]
    \begin{center}
    \centering
    \fboxsep=0mm
    \fboxrule=1pt
    \vspace{.4em}
    \vskip-6pt
    \rotatebox{90}{\begin{minipage}[b]{40pt}{\begin{center} \footnotesize\text{GAN-S} \end{center}}\end{minipage}}\,{\includegraphics[width=0.9\textwidth]
    {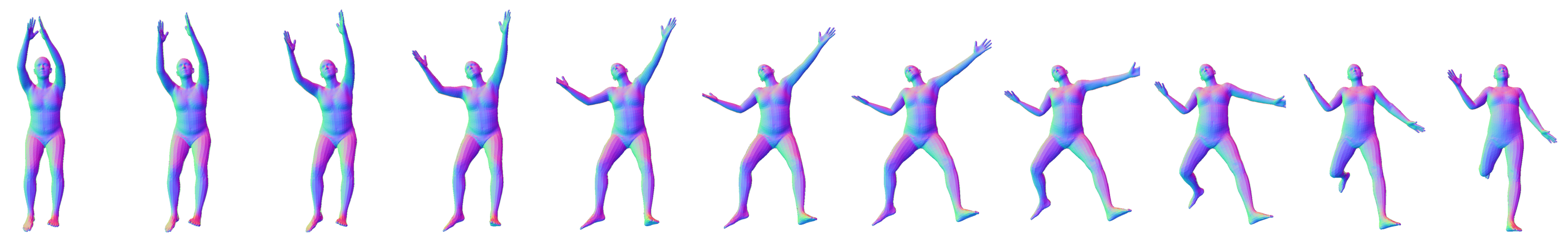}}\\

    \fboxsep=0mm
    \vskip-6pt
    \rotatebox{90}{\begin{minipage}[b]{40pt}{\begin{center} \footnotesize\text{GAN-U} \end{center}}\end{minipage}}\,{\includegraphics[width=0.9\textwidth]
    {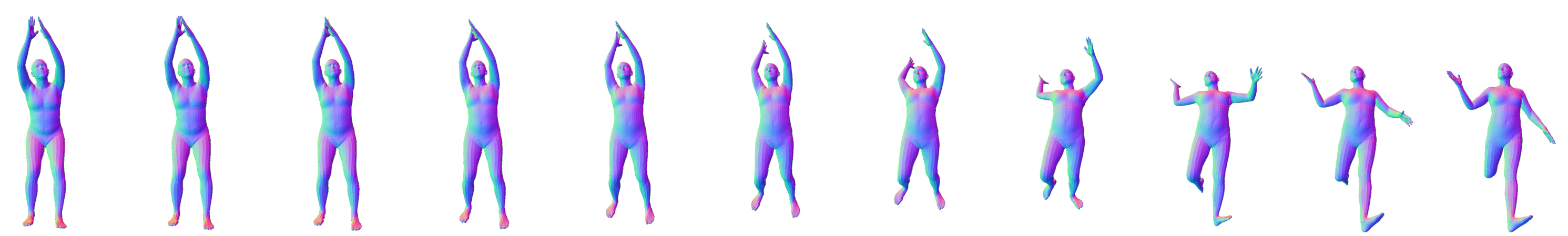}}\\

	\fboxsep=0mm
    \vskip-6pt
    \rotatebox{90}{\begin{minipage}[b]{40pt}{\begin{center} \footnotesize\text{GAN-N} \end{center}}\end{minipage}}\,{\includegraphics[width=0.9\textwidth]
    {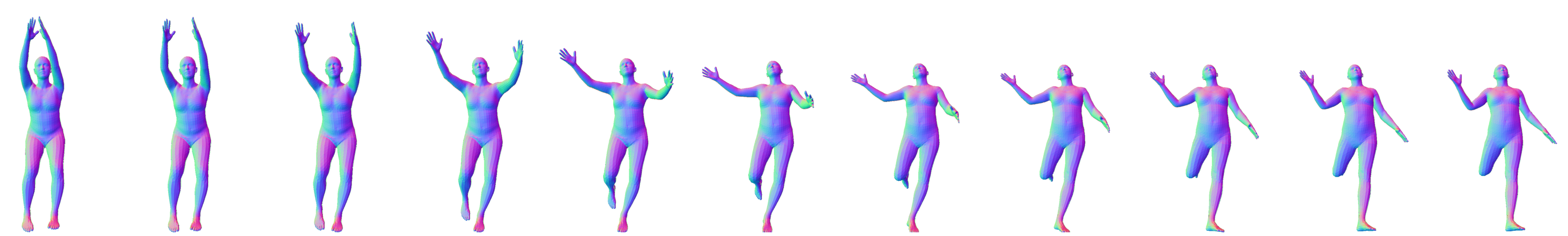}}\\

	\fboxsep=0mm
    \vskip-6pt
    \rotatebox{90}{\begin{minipage}[b]{40pt}{\begin{center} \footnotesize\text{VPoser~\cite{Pavlakos19}} \end{center}}\end{minipage}}\,{\includegraphics[width=0.9\textwidth]
    {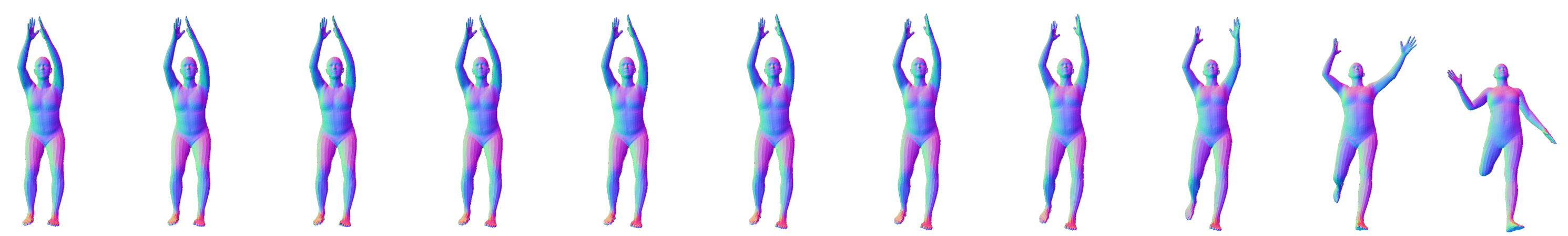}}\\

    \noindent\rule{\linewidth}{0.4pt}
    
    \fboxsep=0mm
    \rotatebox{90}{\begin{minipage}[b]{40pt}{\begin{center} \footnotesize\text{GAN-S} \end{center}}\end{minipage}}\,{\includegraphics[width=0.9\textwidth]
    {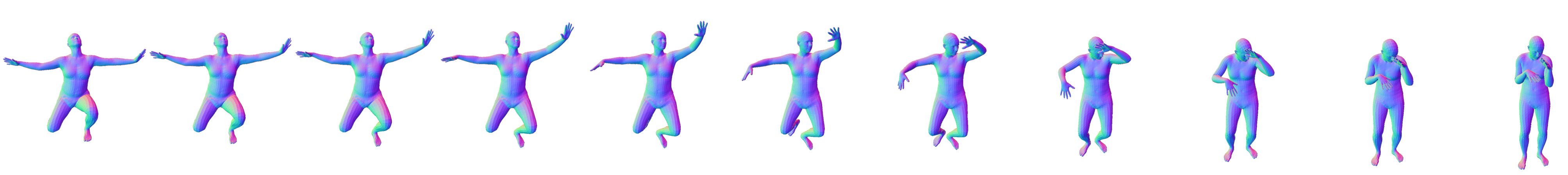}}\\

    \fboxsep=0mm
    \vskip-6pt
    \rotatebox{90}{\begin{minipage}[b]{40pt}{\begin{center} \footnotesize\text{GAN-U} \end{center}}\end{minipage}}\,{\includegraphics[width=0.9\textwidth]
    {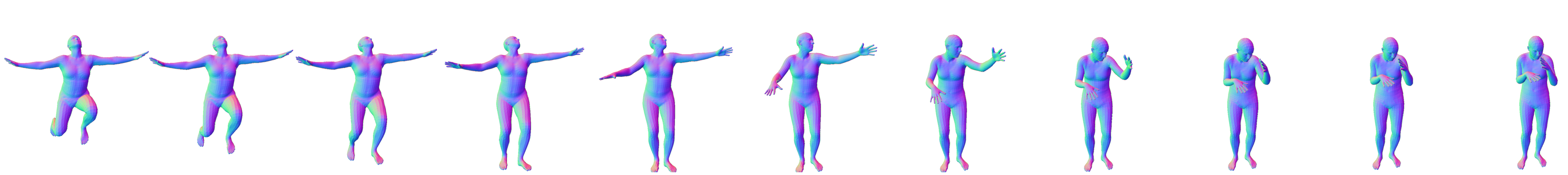}}\\
    
	\fboxsep=0mm
    \vskip-6pt
    \rotatebox{90}{\begin{minipage}[b]{40pt}{\begin{center} \footnotesize\text{GAN-N} \end{center}}\end{minipage}}\,{\includegraphics[width=0.9\textwidth]
    {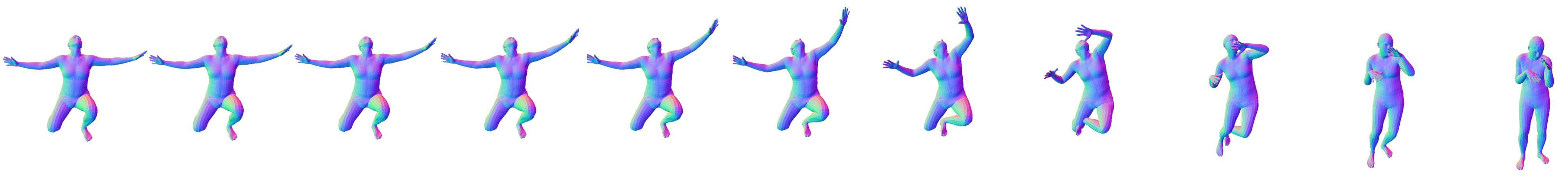}}\\

	\fboxsep=0mm
    \vskip-6pt
    \rotatebox{90}{\begin{minipage}[b]{40pt}{\begin{center} \footnotesize\text{VPoser~\cite{Pavlakos19}} \end{center}}\end{minipage}}\,{\includegraphics[width=0.9\textwidth]
    {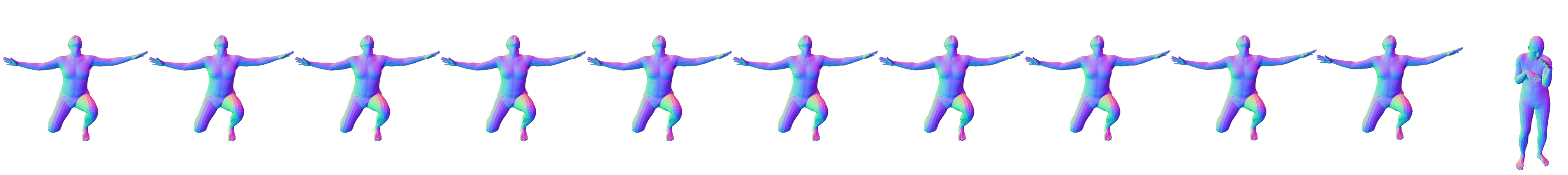}}\\

    \end{center}
    \vskip-6pt
        \caption{Examples of interpolations for different generative models. All GAN models provide smooth (yet semantically very different) interpolations, while VPoser~\cite{Pavlakos19} sticks at one pose for most of the path and ``jumps" into the ending pose. More examples can be found in the supplementary material.
    }
    \label{fig:interp_render}
\end{figure*}


We illustrate the behavior of different models in Fig.~\ref{fig:interpolation}, where we average the consecutive interpolation distances $\bar{\Delta}_{ij}$ across all pairs. The closer the curve is to being horizontal, the smoother is the interpolation.
The VPoser~\cite{Pavlakos19} curve indicates that interpolation with this model is subject to jumping from $\bod^{i}_0$ to $\bod^{i}_T$ in very few steps, either at the beginning or the end of the sequence, and the remaining steps are spent performing small pose adjustments.
This effect can also be seen in GAN-N but to a lesser degree.
In contrast, GAN-S and GAN-U both produce smooth interpolations, with a slight advantage to the spherical distribution.
More plots with interpolation distances for particular pairs can be found in the supplementary material. 

To measure the smoothness of an interpolation, we compute the ratio between the maximal and minimal transitions in a sequence $\Delta_{ij}$, i.e.,
\begin{align}
    R_{ij} = \frac{ \max(\Delta_{ij}) } { \min(\Delta_{ij}) } \; .
    \label{eq:interp_norm}
\end{align}
We report the mean, variance and median of the ratios $R_{ij}$ in Table~\ref{table:interp}. These results confirm our previous conclusions: GAN-S yields the smoothest interpolations, closely followed by GAN-U.
In Fig.~\ref{fig:interp_render} we show interpolation between two pairs of samples for different models.


\subsection{Mesh Optimization from 2D Joints}
\label{subsec:optim}
Let us now turn to the optimization application discussed in Section~\ref{sec:applications} and in Fig.~\ref{fig:overview} $(C)$.
Given the 2D joint locations, the optimal latent vector can be found using an iterative optimization algorithms. In our experiments, we use L-BFGS-B for all models. 
In the case of GAN-S and GAN-U, we renormalize the estimated $\bz$ given their input bound at each optimization step. Note that projecting the 3D body joints to the observed 2D joints on the image via the camera project $\Pi$ requires access to the camera parameters and the body orientations. We obtain them in the same way as in~\cite{Bogo16}.


\begin{table}[t]
	
	\caption{Reporting 3D pose error for the bodies recovered through optimization from 2D joint targets. (numbers in \textit{mm}).}
	\vspace{-4mm}
	\begin{small}
	\begin{center}
			\begin{tabular}{c|c}
				 & P-MPJPE ($\downarrow$) \\
          \hline
				GAN-S (\ours{}) &  \textbf{84.3}\\
				GAN-U (\ours{}) &  90.7\\
				GAN-N (\ours{}) &  85.4\\
				VPoser~\cite{Pavlakos19} 			&  90.1 \\
				GMM ~\cite{Bogo16} & 92.3

		\end{tabular}
	\end{center}
	\end{small}
	\vspace{-6mm}
	\label{table:optim}
\end{table}

In Table~\ref{table:optim}, we report the 3D pose errors (after rigid alignment) obtained by recovering the SMPL parameters $\hat{\Theta}$ from 2D joints for the H3.6M dataset~\cite{Ionescu14a}, following \textit{Protocol 2}. Note that GAN-S again yields the best results for this application, this time closely followed by GAN-N. By contrast, GAN-U yields a higher error, indicating its input bounds makes it less suitable for the optimization-based tasks.


\subsection{Image to Mesh Regression}
\label{sec:imageToMesh}


\begin{table}[t]

	\caption{Reporting 3D pose error for the bodies recovered through regression from input images.
		(numbers in \textit{mm}).}
	\vspace{-4mm}
	\begin{small}
	\begin{center}
			\begin{tabular}{c|cc}
				  & MPJPE ($\downarrow$) & P-MPJPE ($\downarrow$) \\
          \hline
        $\text{HMR}_{\text{[COCO EFT]}}$~\cite{Joo20} & 75.91 & 68.02 \\
		\hline
				GAN-S (\ours{}) & \textbf{63.24} & \textbf{56.32} \\
				GAN-U (\ours{}) & 64.28 		 & 57.30 \\
				GAN-N (\ours{}) & 67.19 		 & 59.92 \\
				VPoser~\cite{Pavlakos19}			& 69.18 		 & 61.71 
		\end{tabular}
	\end{center}
	\end{small}
	\vspace{-8mm}
	\label{table:hmr}
\end{table}

We train the approach to body regression from an image introduced in Section~\ref{sec:applications} and in Fig.~\ref{fig:overview} $(D)$ for the three versions of our approach and for the VPoser~\cite{Pavlakos19}.
We report accuracy results on test data in Table~\ref{table:hmr} in terms of 3D pose error of the recovered bodies after Procrustes alignment (P-MPJPE), according to \textit{Protocol 2} of ~\cite{Ionescu14a}. Once again GAN-S performs the best, with GAN-U and GAN-N outperforming VPoser. Our models also report better accuracy compared to \cite{Joo20}, for which the accuracy of the pre-trained model, provided by the authors, is reported.


\section{Limitations}
\label{sec:limitations}

Using our GAN priors, the diversity of their learned distributions is limited by their training sets, which might not be diverse enough for downstream tasks. This is, however, similar to any other model that learns a distribution such as VPoser or HMR, which is also limited by its training distribution.


\section{Conclusion}
\label{sec:conclusion}
In this paper we proposed a simple yet effective prior for SMPL model to bound it to realistic human poses.
We show that the learned prior can cover the diversity of the training distribution, while also being capable of generating novel unseen samples. Further, we demonstrate the advantage of learning such a prior in generation, optimization, and regression based frameworks, where the learned prior can be trained once and for all, then used in any downstream task without requiring to balance different losses. Our results show that using a spherical distribution for the learned prior leads to smoother transition in the generated samples from the latent space, while also yielding more accurate results for optimization- and regression-based tasks, indicating this prior is better suited for learning human poses.

{\small
\bibliographystyle{ieee_fullname}
\bibliography{string,vision,learning,local_bib}
}

\clearpage


\twocolumn[{%
 \centering
 \bf\Large Adversarial Parametric Pose Prior\\Supplementary Material\\\quad
}]

\appendix

\section{Proof of SLERP formula}

Let us denote two vectors of unit length:

\begin{align}
    \begin{cases}
        &\bx,\by \in \mathbb{S}^d\\
        &\|\bx\|_2=\|\by\|_2=1\\
        &<\bx\cdot \by>=\cos\theta, \quad\theta \in (0,\pi)
    \end{cases}
    \label{eq:xy}
\end{align}\;

where $<\bx \cdot \by>$ is the inner product between vector $\bx$ and $\by$, with $\theta$ as the angle between them on a $d$-dimensional unit sphere $\mathbb{S}^d$. With slight abuse of notations, we use $\bx$ to refer to both a point on a sphere, when talking about the sampled $d$-dimensional point, and also a vector from the origin of this sphere, when talking about  trigonometric operations.

The point $\bz \in \mathbb{S}^d$ that lies on the interpolation path from $\bx$ to $\by$ can be found via the Spherical Linear Interpolation (SLERP):

\begin{align}
    \bz = \frac{\sin(\theta-\phi)}{\sin{\theta}} \bx + \frac{\sin\phi}{\sin{\theta}} \by
    \label{eq:slerp_formula}
\end{align}\,

where

\begin{align}
    \begin{cases}
        &\phi=\arccos(<\bz\cdot \bx>), \quad\phi \in [0,\theta]\\
        &\|\bz\|_2 = 1
    \end{cases}
    \label{eq:z_facts}
\end{align}

Initially, SLERP arose in the task of 3D rotations for solid objects~\cite{Shoemake85} and Eq.~\ref{eq:slerp_formula} can be naturally derived from the 4D rotation intuition via quaternions.
However, Eq.~\ref{eq:slerp_formula} is still valid for $d$-dimensional vectors, even though quaternion intuition is not applicable anymore. Next, we prove interpretation used in this equation.
\newline

\underline{\textbf{Proof}}

We want to find the point $\bz$ on the sphere $\mathbb{S}^d$ that is a linear combination of $\bx$ and $\by$:

\begin{align}
    \bz = \alpha\bx + \beta\by
    \label{eq:linear_form}
\end{align}

First, we derive the connection between coefficients $\alpha$ and $\beta$ in Eq.~\ref{eq:linear_form}. Knowing each point should lie on the unit-sphere we have:
\begin{align}
    1
    &= \|\bz\|^2 \nonumber\\
    &= \|\alpha\bx + \beta\by\|^2\nonumber\\
    &= \alpha^2 + \beta^2 + 2\alpha\beta\cos\theta\nonumber\\
    &= (\alpha^2 + 2\alpha\beta\cos\theta + \beta^2\cos^2\theta) + \beta^2 - \beta^2\cos^2\theta\nonumber\\
    &= (\alpha + \beta\cos\theta)^2 + (\beta\sin\theta)^2\nonumber\\
    &=\tilde{\alpha}^2 + \tilde{\beta}^2\label{eq:alpha_beta_link}\,
\end{align}

where

\begin{align}
    \begin{cases}
        &\tilde{\alpha} = \alpha + \beta \cos\theta \\
        &\tilde{\beta} = \beta \sin\theta
    \end{cases}
    \label{eq:tilde_coefs}
\end{align}

Then, we rewrite the denotion for $\phi$ from Eq.~\ref{eq:z_facts} and obtain the value of $\tilde{\alpha}$:
\begin{align}
    \cos\phi
    &= <\bz\cdot \bx>\nonumber\\
    &=<(\alpha \bx + \beta \by) \cdot \bx>\nonumber\\
    &=\alpha + \beta \cos\theta\nonumber\\
    &=\tilde{\alpha}
    \label{eq:tilde_alpha}
\end{align}

Given $\tilde{\alpha}$ and $\tilde{\beta}$ formulations in Eqs.~\ref{eq:alpha_beta_link} and \ref{eq:tilde_alpha}, one can derive $\tilde{\beta} = \sin{\phi}$. Finally, to obtain the coefficients $\alpha$ and $\beta$ in Eq.~\ref{eq:linear_form}, we use Eqs.~\ref{eq:tilde_coefs} and \ref{eq:tilde_alpha} as follows:

\begin{align}
    \begin{cases}
    &\beta = \frac{\tilde{\beta}}{\sin{\theta}} = \frac{\sin{\phi}}{\sin{\theta}}\\
    &\alpha = \tilde{\alpha} - \beta\cos\theta = \cos\phi - \frac{\sin{\phi}}{\sin{\theta}}\cos\theta = \frac{\sin{(\theta-\phi)}}{\sin{\theta}} ,
    \end{cases}
\end{align}\,

using which we can derive Eq.~\ref{eq:slerp_formula}.
Choosing $\phi = \frac{t\theta}{T}$, where $t \in [0,T]$, one gets the desired interpolation between two points $\bx$ and $\by$, as presented in Eq.~\ref{eq:slerp}, which is linear in parameter $t$.

\section{Interpolation sequences}


\begin{figure*}[t]
    \begin{center}
    \centering
    \fboxsep=0mm
    \fboxrule=1pt
    \vspace{.4em}
    \vskip-6pt
    \rotatebox{90}{\begin{minipage}[b]{40pt}{\begin{center} \footnotesize\text{\quad GAN-S} \end{center}}\end{minipage}}\,{\includegraphics[width=0.8\textwidth]
    {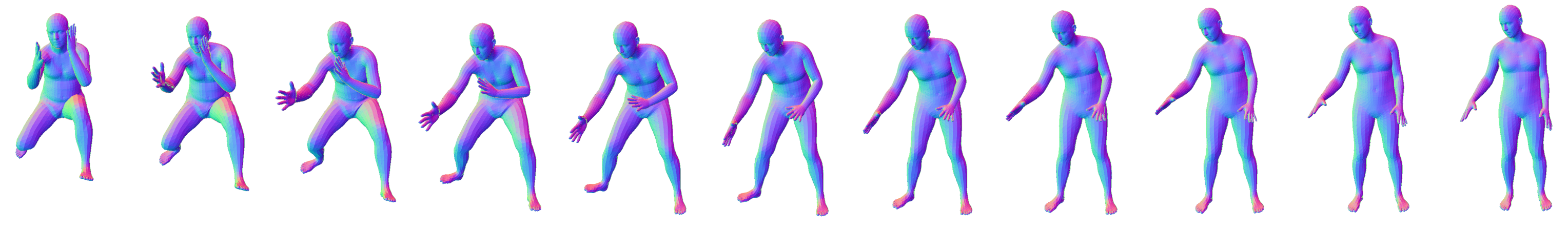}}\\

    \fboxsep=0mm
    \vskip-6pt
    \rotatebox{90}{\begin{minipage}[b]{40pt}{\begin{center} \footnotesize\text{\quad GAN-U} \end{center}}\end{minipage}}\,{\includegraphics[width=0.8\textwidth]
    {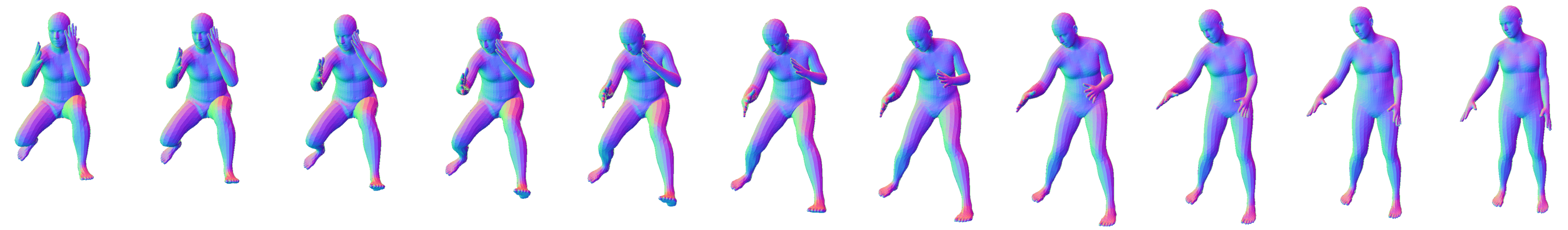}}\\

	\fboxsep=0mm
    \vskip-6pt
    \rotatebox{90}{\begin{minipage}[b]{40pt}{\begin{center} \footnotesize\text{\quad GAN-N} \end{center}}\end{minipage}}\,{\includegraphics[width=0.8\textwidth]
    {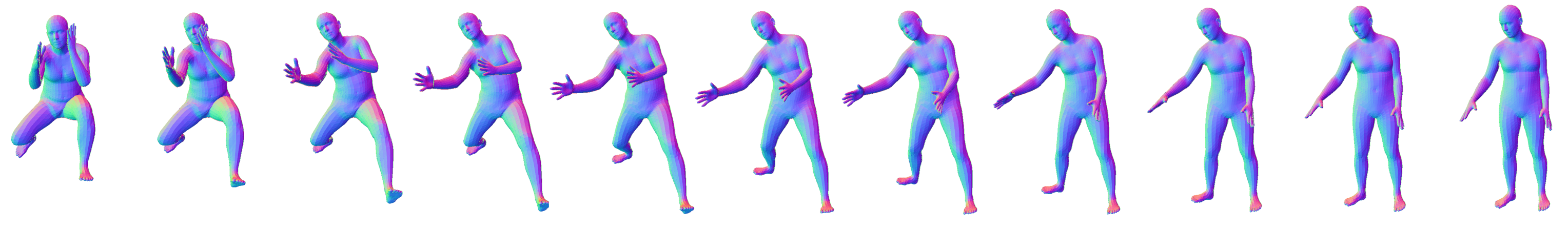}}\\

	\fboxsep=0mm
    \vskip-6pt
    \rotatebox{90}{\begin{minipage}[b]{40pt}{\begin{center} \footnotesize\text{\quad VPoser~\cite{Pavlakos19}} \end{center}}\end{minipage}}\,{\includegraphics[width=0.8\textwidth]
    {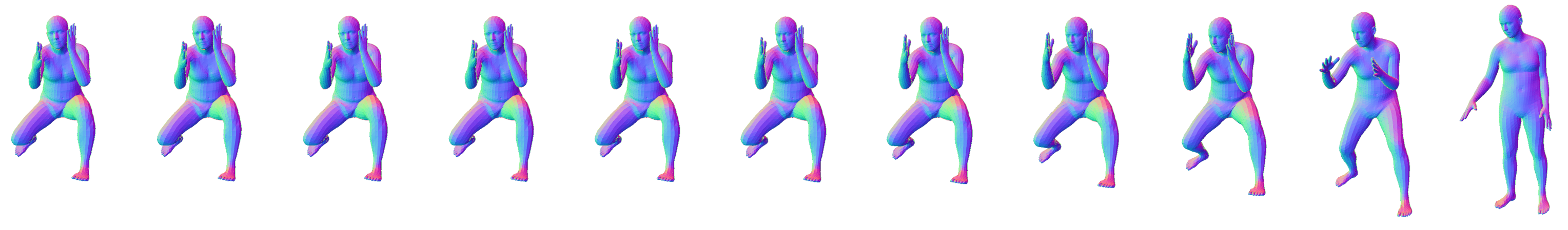}}\\

    \vskip-8pt
    \noindent\rule{0.8\linewidth}{0.4pt}

    \fboxsep=0mm
    \rotatebox{90}{\begin{minipage}[b]{40pt}{\begin{center} \footnotesize\text{\quad GAN-S} \end{center}}\end{minipage}}\,{\includegraphics[width=0.8\textwidth]
    {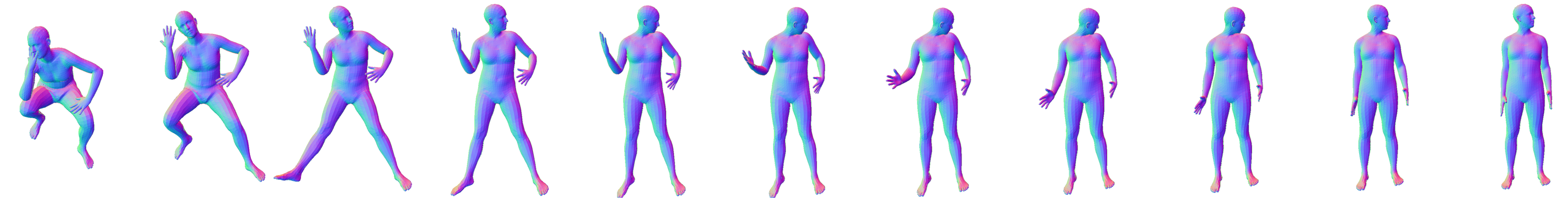}}\\

    \fboxsep=0mm
    \vskip-6pt
    \rotatebox{90}{\begin{minipage}[b]{40pt}{\begin{center} \footnotesize\text{\quad GAN-U} \end{center}}\end{minipage}}\,{\includegraphics[width=0.8\textwidth]
    {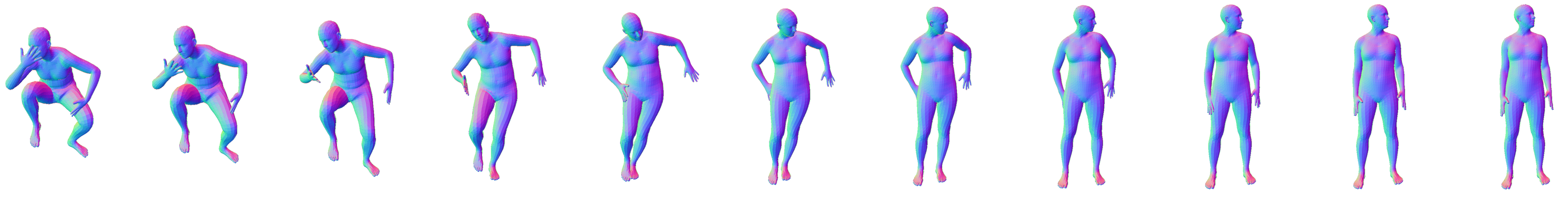}}\\

	\fboxsep=0mm
    \vskip-6pt
    \rotatebox{90}{\begin{minipage}[b]{40pt}{\begin{center} \footnotesize\text{\quad GAN-N} \end{center}}\end{minipage}}\,{\includegraphics[width=0.8\textwidth]
    {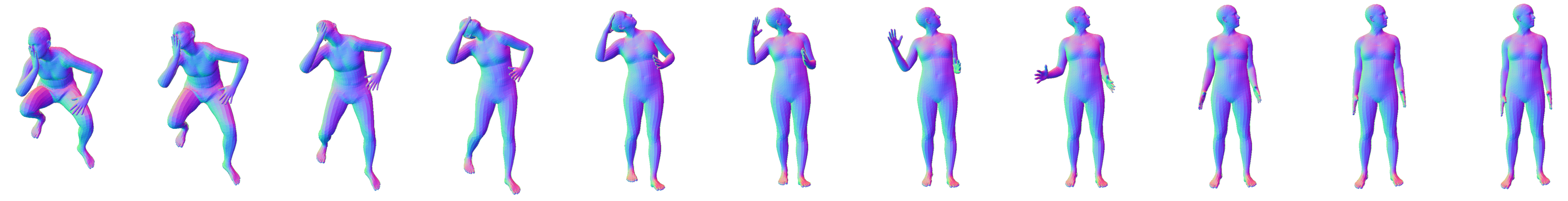}}\\

	\fboxsep=0mm
    \vskip-6pt
    \rotatebox{90}{\begin{minipage}[b]{40pt}{\begin{center} \footnotesize\text{\quad VPoser~\cite{Pavlakos19}} \end{center}}\end{minipage}}\,{\includegraphics[width=0.8\textwidth]
    {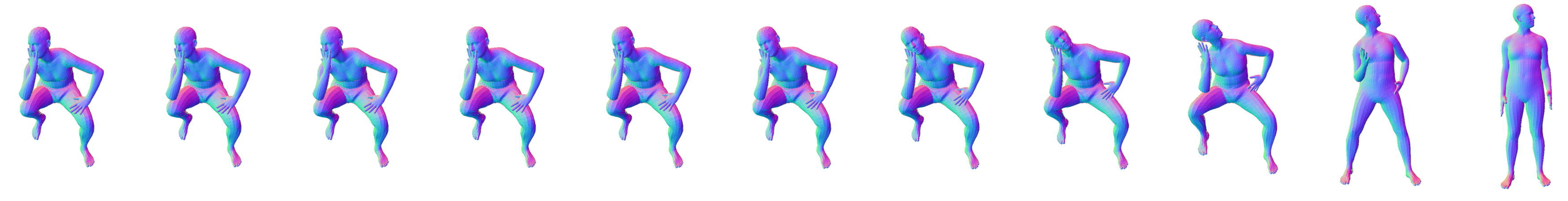}}\\
    
    \vskip-8pt
    \noindent\rule{0.8\linewidth}{0.4pt}

    \fboxsep=0mm
    \rotatebox{90}{\begin{minipage}[b]{40pt}{\begin{center} \footnotesize\text{GAN-S} \end{center}}\end{minipage}}\,{\includegraphics[width=0.8\textwidth]
    {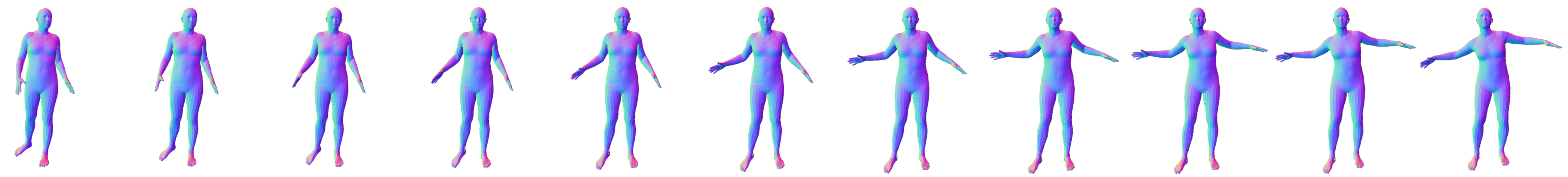}}\\

    \fboxsep=0mm
    \vskip-6pt
    \rotatebox{90}{\begin{minipage}[b]{40pt}{\begin{center} \footnotesize\text{GAN-U} \end{center}}\end{minipage}}\,{\includegraphics[width=0.8\textwidth]
    {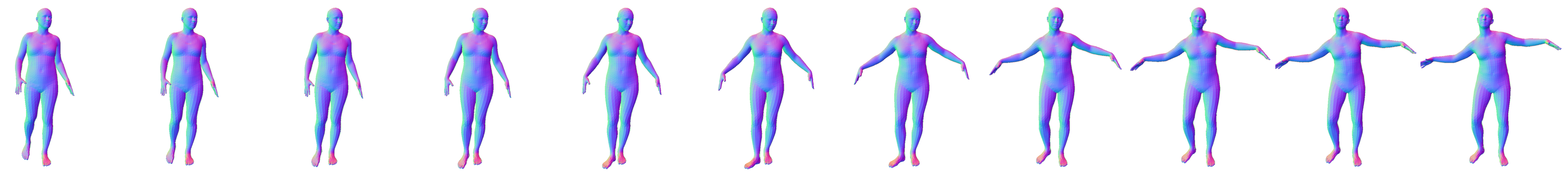}}\\

	\fboxsep=0mm
    \vskip-6pt
    \rotatebox{90}{\begin{minipage}[b]{40pt}{\begin{center} \footnotesize\text{GAN-N} \end{center}}\end{minipage}}\,{\includegraphics[width=0.8\textwidth]
    {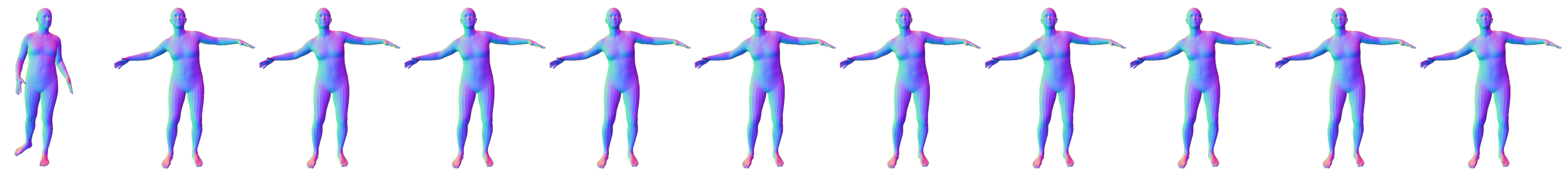}}\\

	\fboxsep=0mm
    \vskip-6pt
    \rotatebox{90}{\begin{minipage}[b]{40pt}{\begin{center} \footnotesize\text{VPoser~\cite{Pavlakos19}} \end{center}}\end{minipage}}\,{\includegraphics[width=0.8\textwidth]
    {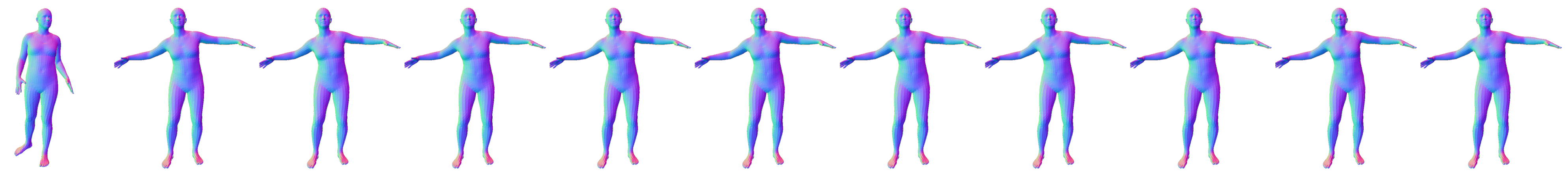}}\\

    \end{center}
    \vskip-6pt
        \caption{Examples of interpolations for different generative models, 10 intermediate interpolants each. Corresponding transition distances (100-step interpolations) are provided in Fig.~\ref{fig:supp_interp_plots}.
    }
    \label{fig:supp_interp_render}
\end{figure*}


\begin{figure*}[t]%
    \centering
    \subfloat[Transition distances for the sequence in Fig.~\ref{fig:supp_interp_render} \textit{top}.\\
    Smooth for GANs, abrupt for VAE.
    ]{{\includegraphics[width=0.32\textwidth]{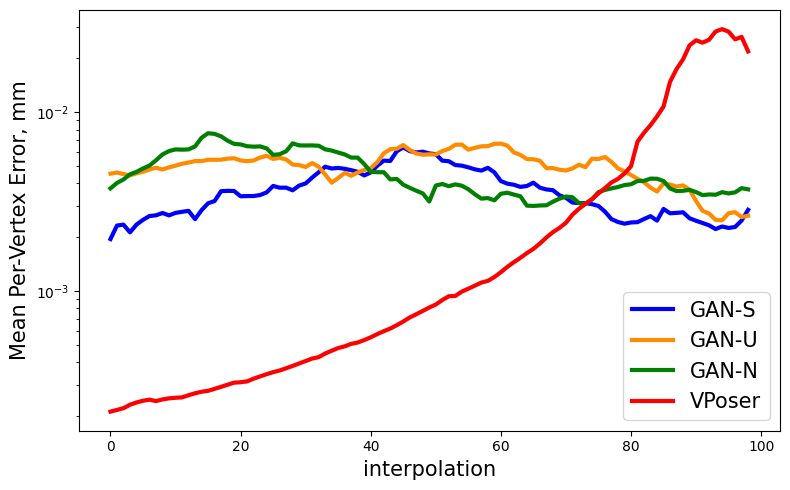} }}%
    \hspace{0.01cm}
    \subfloat[Transition distances for the sequence in Fig.~\ref{fig:supp_interp_render} \textit{middle}.\\
    All GANs provide hardly smooth yet physically plausible interpolation, while VPoser~\cite{Pavlakos19} smoothly moves the \textit{head} of the body and then "jumps" to the final pose.
    ]{{\includegraphics[width=0.32\textwidth]{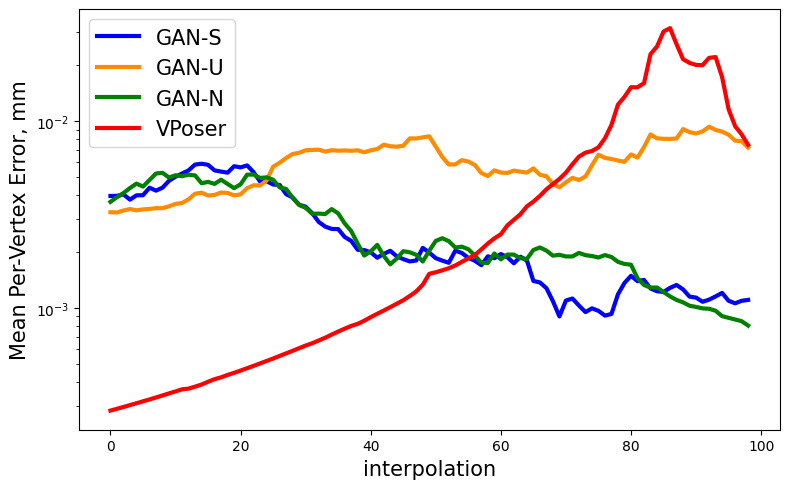} }}%
    \hspace{0.01cm}
    \subfloat[Transition distances for the sequence in Fig.~\ref{fig:supp_interp_render} \textit{bottom}.\\
    GAN-N model demonstrates less smooth interpolations than bounded GAN models, closer to VPoser~\cite{Pavlakos19} and locally sticks at one pose (transiton distances $\sim 10^{-5}$ mm per vertex).
    ]{{\includegraphics[width=0.32\textwidth]{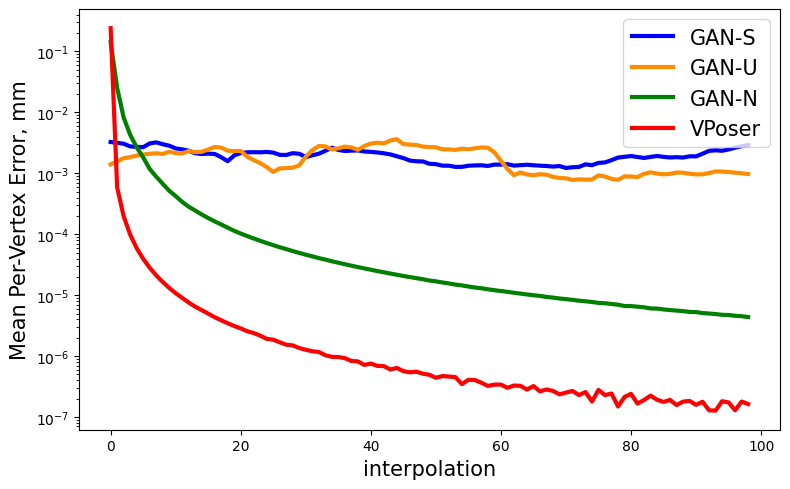} }}%
    \hspace{0.01cm}
    \caption{Mean per-vertex transition distances of different generative models for the samples presented in Fig.~\ref{fig:supp_interp_render}. 
    Each interpolation path takes 100 steps from start to end, while the corresponding images in Fig.~\ref{fig:supp_interp_render} show only each 10-th step.
    It is clear that VAE-based VPoser~\cite{Pavlakos19} provides abrupt transitions, while GAN-based models provide smoother transitions in the output space.
    }%
    \label{fig:supp_interp_plots}%
\end{figure*}

In Fig.~\ref{fig:supp_interp_render} we show more interpolation sequences for different generative models: GAN-based GAN-S, GAN-U, GAN-N and VAE-based VPoser~\cite{Pavlakos19}. The procedure of sampling and interpolation is the same as in the manuscript. Also, in Fig.~\ref{fig:supp_interp_plots} we provide the transition distances that correspond to the interpolations in Fig.~\ref{fig:supp_interp_render}.
For GAN-based models transition distances for the sampled pairs lie in the range [$10^{-3}$, $10^{-2}$] mm (per-vertex distance), which is in much smaller range than VPoser. Note that the average result for all samples is provided in the main paper (Fig.~\ref{fig:interpolation}).
It is clear that VAE-based VPoser~\cite{Pavlakos19} applies most of the transition either at the beginning or at the end of interpolations, while GAN-based models, especially GAN-S and GAN-U, provide smooth interpolations.

\section{Image-to-Mesh regression}

\begin{figure*}[t]

    \begin{center}

    {\includegraphics[width=0.5\textwidth]
    {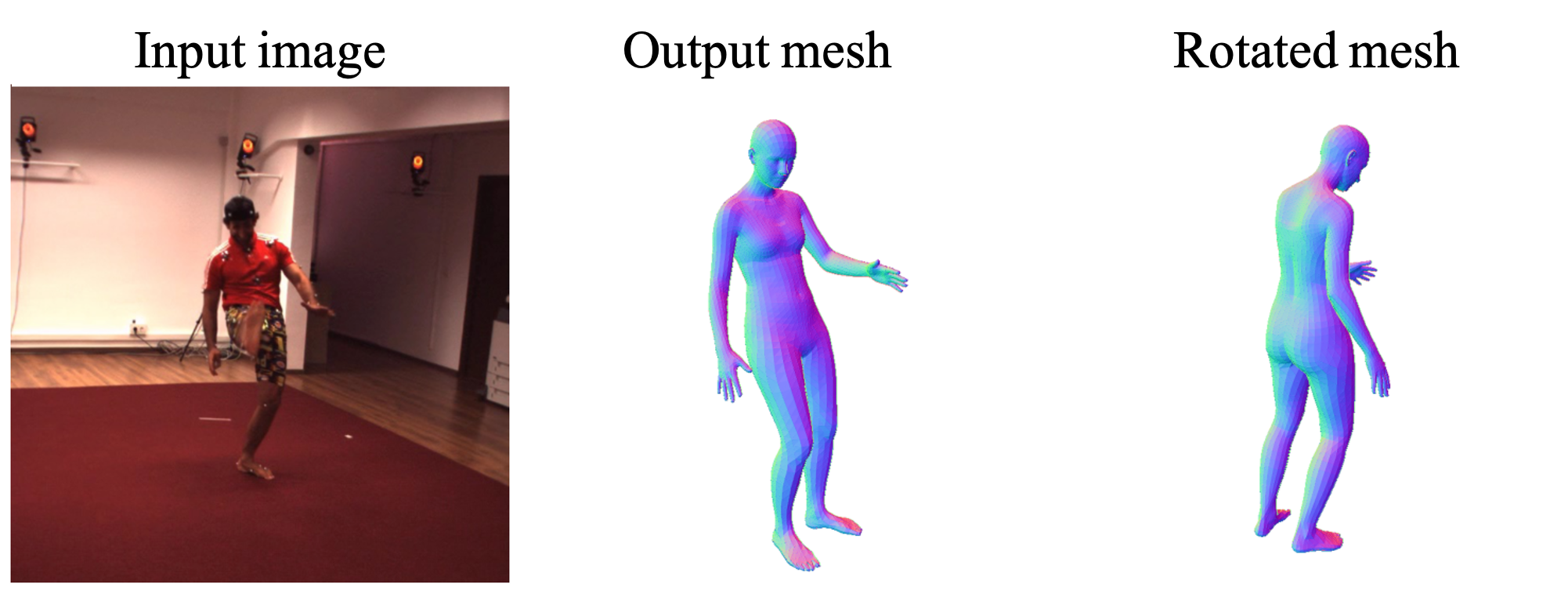}}\hfill
    {\includegraphics[width=0.5\textwidth]
    {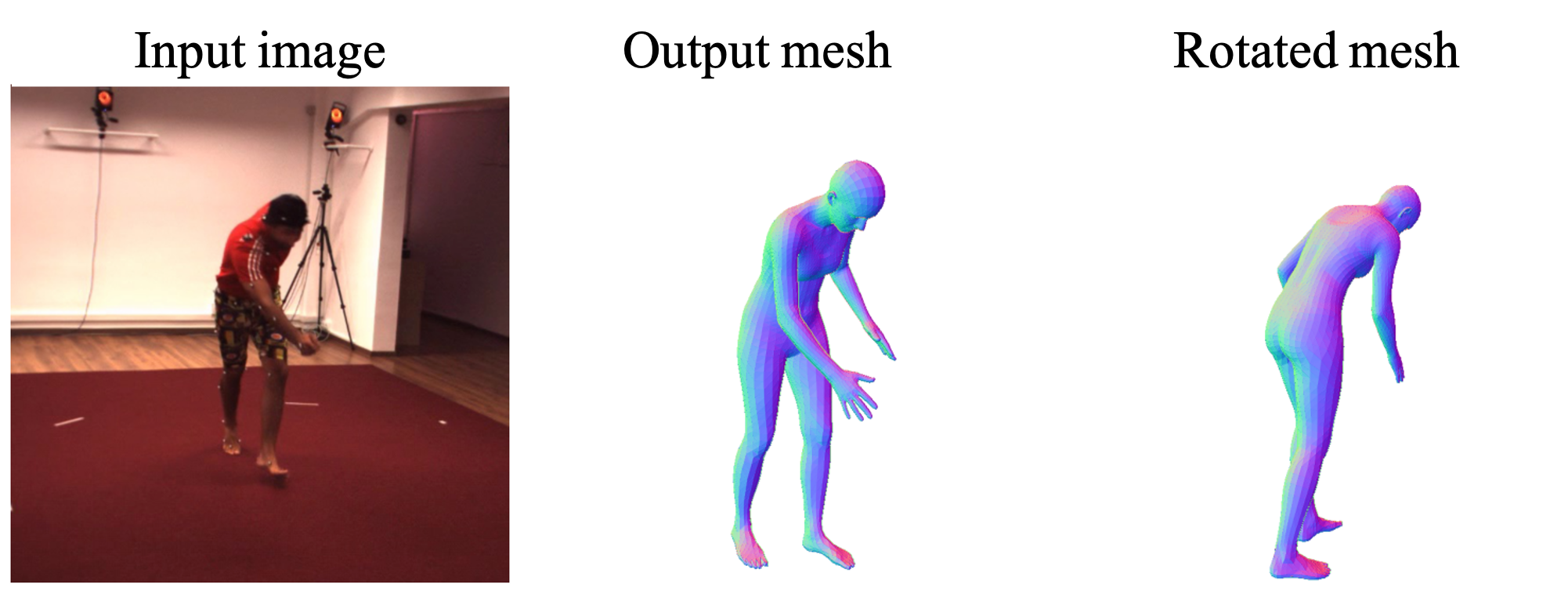}}\\

    {\includegraphics[width=0.5\textwidth]
    {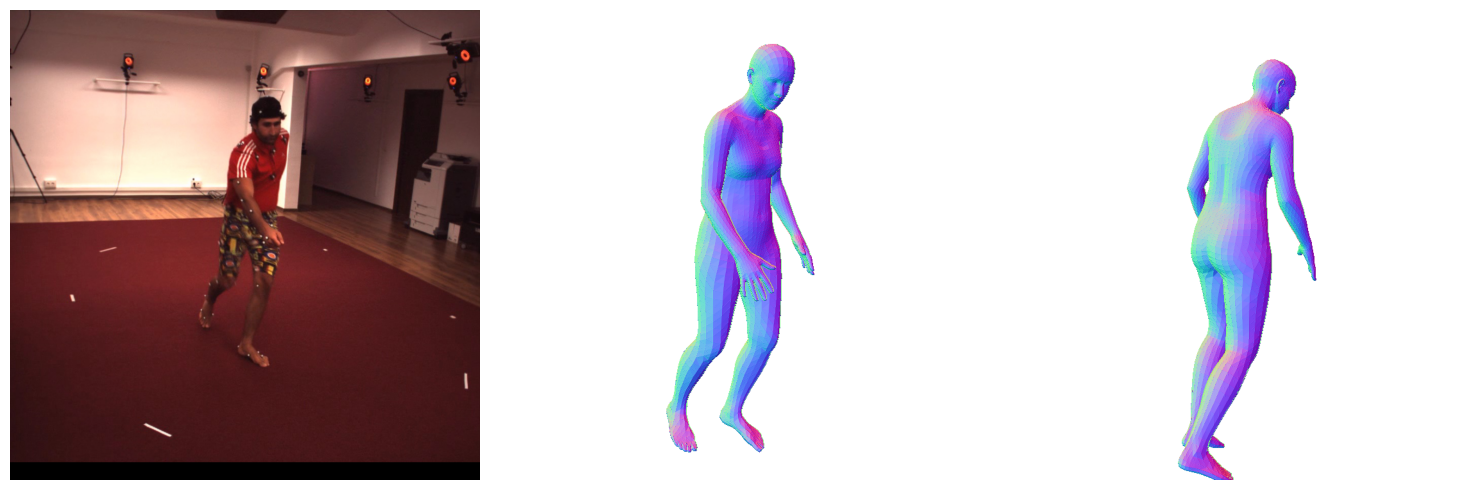}}\hfill
    {\includegraphics[width=0.5\textwidth]
    {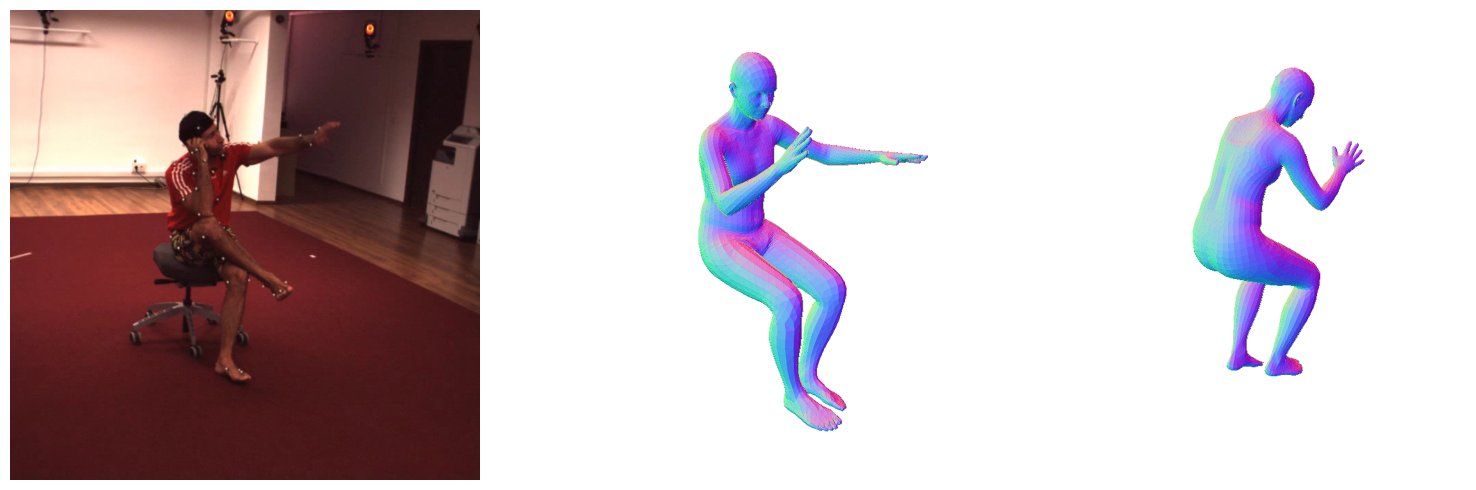}}\\

    {\includegraphics[width=0.5\textwidth]
    {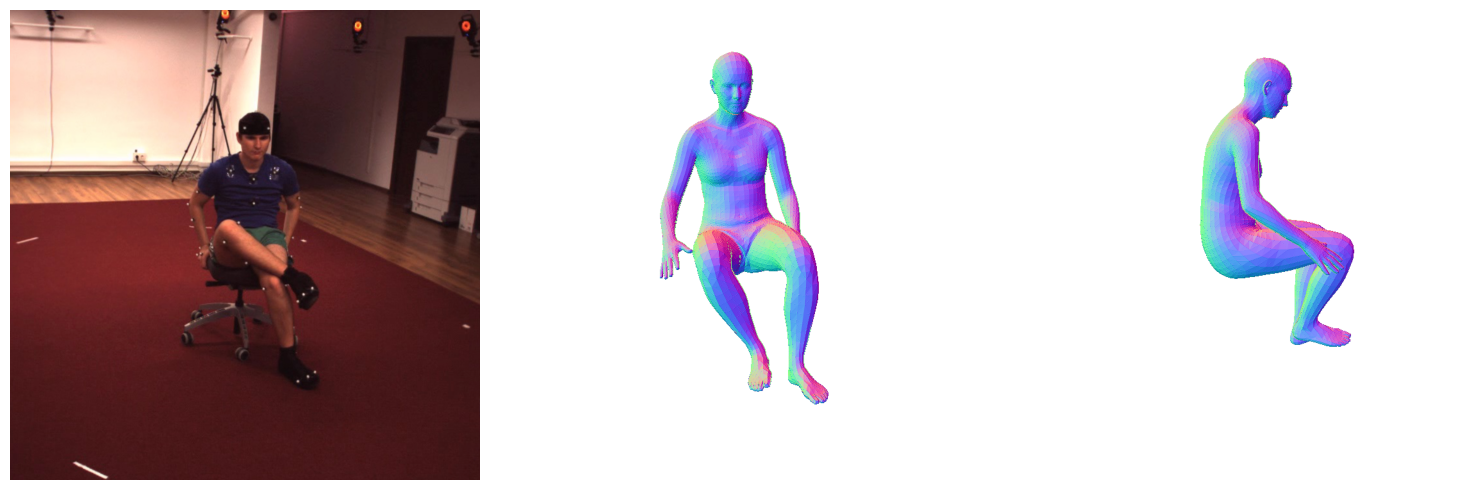}}\hfill
    {\includegraphics[width=0.5\textwidth]
    {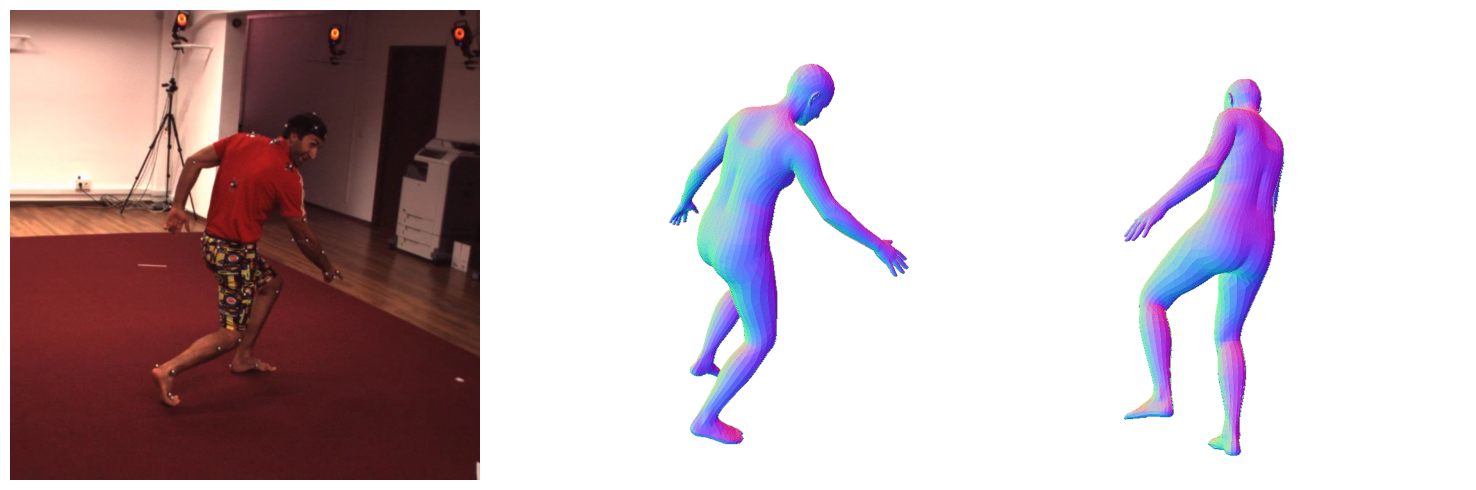}}\\

    {\includegraphics[width=0.5\textwidth]
    {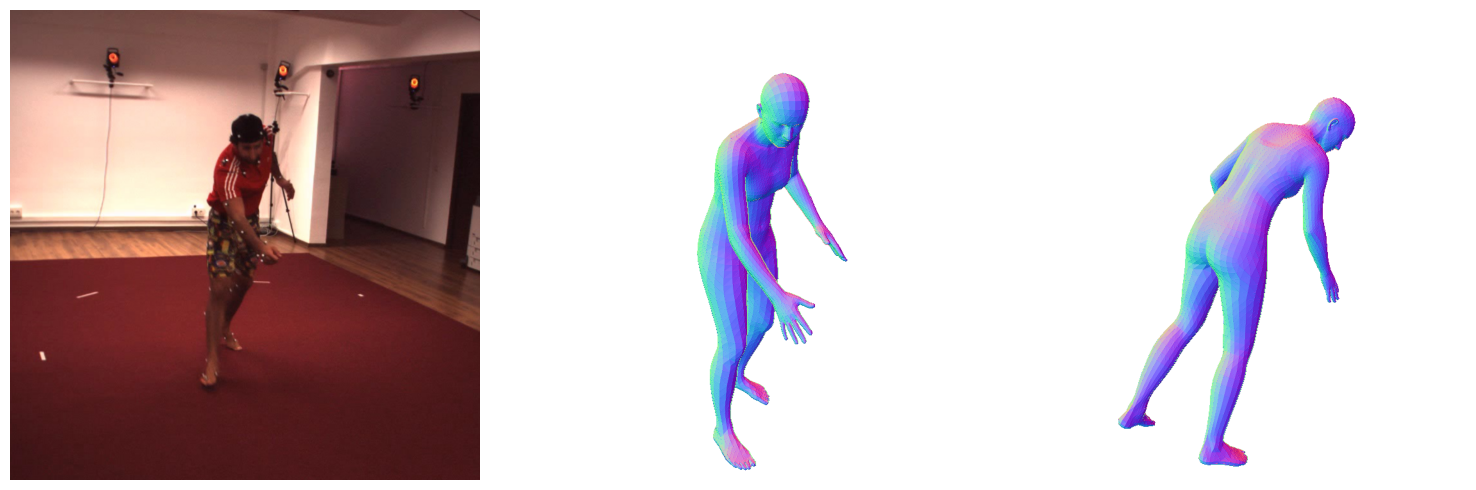}}\hfill
    {\includegraphics[width=0.5\textwidth]
    {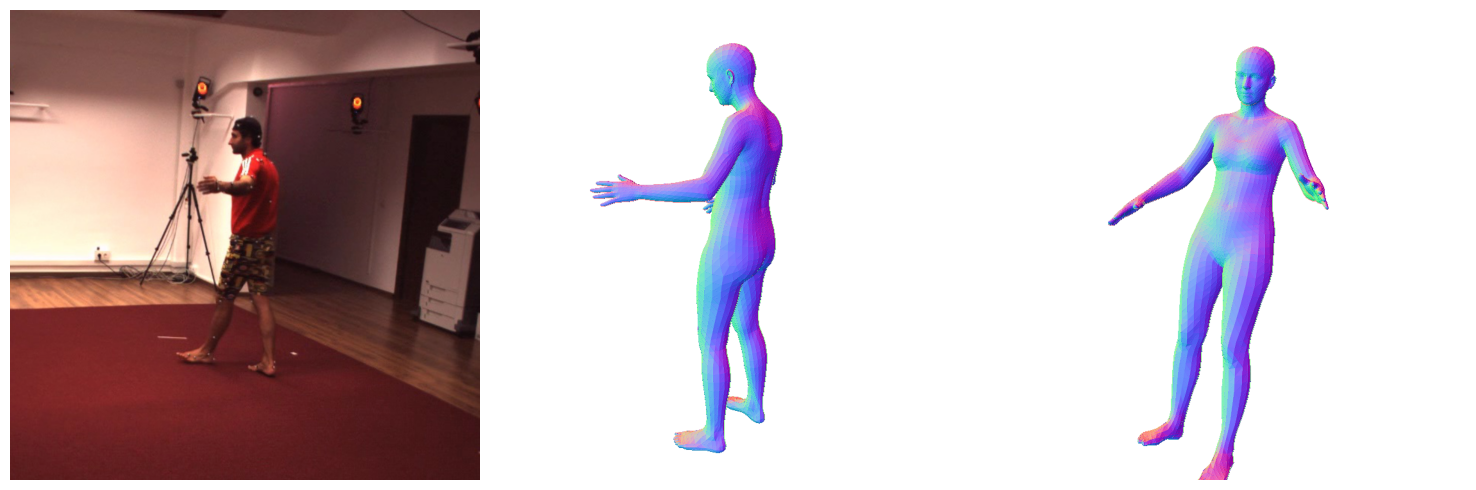}}\\

    \end{center}
    \vskip-6pt
        \caption{
            Mesh predictions in Section~\ref{sec:imageToMesh} experiments of the main paper with the highest (``worst") P-MPJPE on the subset of H3.6M~\cite{Ionescu14a} validation set (according to H3.6M \textit{Protocol 2}). 
            Examples are split in triplets: \textit{left} - the input image, \textit{middle} - the predicted mesh (with predicted rotation), \textit{right} - the same mesh rotated by $90^{\circ}$ along vertical axis. 
            The error decreases \textit{from left to right} and \textit{top to bottom} (\textit{top-left} sample obtains the highest P-MPJPE value). 
    }
    \label{fig:i2m_h36m}
\end{figure*}

We demonstrate qualitative results of injecting our module GAN-S into the pretrained HMR architecture and fine-tuning it on recent in-the-wild pseudo ground-truth dataset $\text{COCO}_{\text{EFT}}$~\cite{Joo20} (see Sections ~\ref{sec:applications} and ~\ref{sec:imageToMesh} in the main paper for details). We show the worst predictions of our model on H3.6M~\cite{Ionescu14a} validation subset, chosen according to \textit{Protocol 2} (corresponding to the results in Table~\ref{table:hmr} of the main paper).

\section{GANs for shape parameters}


\begin{figure*}[t]
    \begin{center}
    {\includegraphics[width=0.8\textwidth]{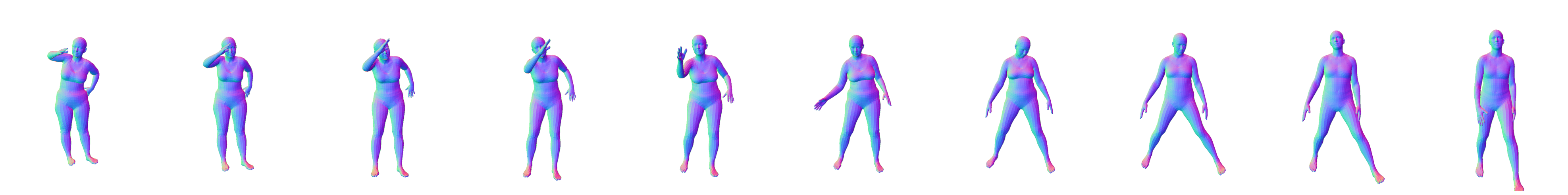}}\\
    {\includegraphics[width=0.8\textwidth]{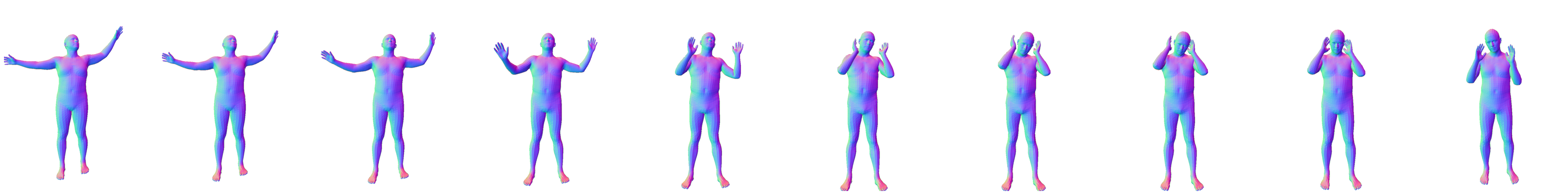}}\\
    {\includegraphics[width=0.8\textwidth]{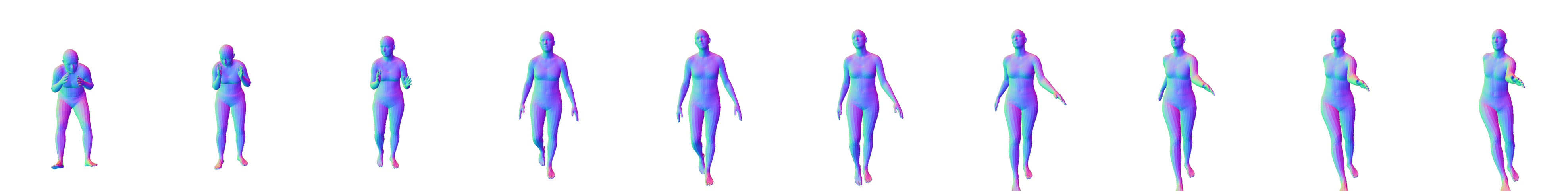}}\\
    {\includegraphics[width=0.8\textwidth]{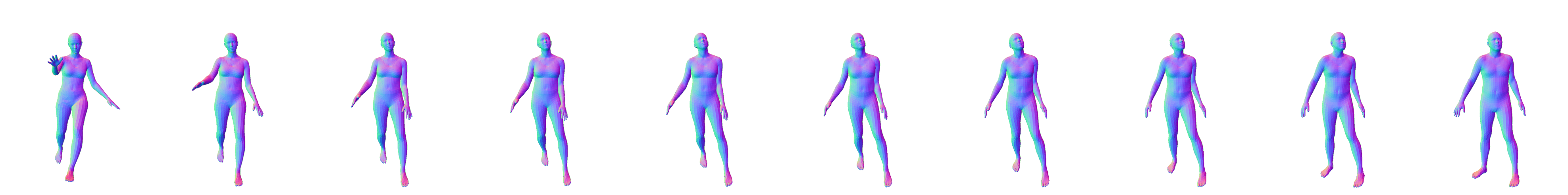}}\\
    {\includegraphics[width=0.8\textwidth]{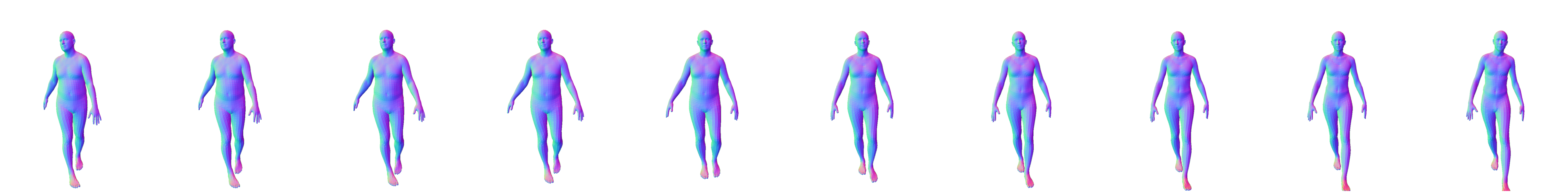}}\\
    {\includegraphics[width=0.8\textwidth]{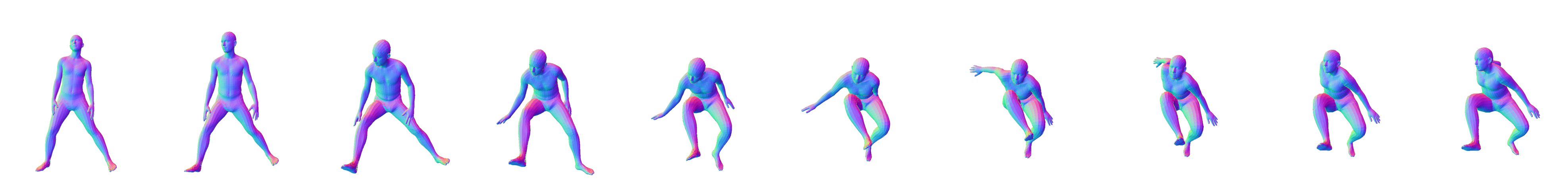}}\\
    {\includegraphics[width=0.8\textwidth]{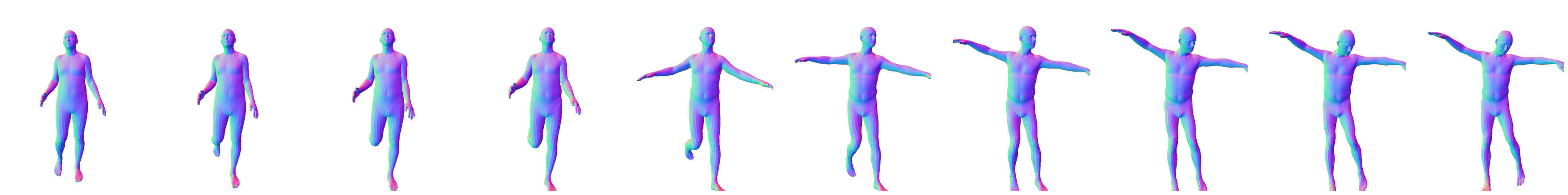}}\\
    {\includegraphics[width=0.8\textwidth]{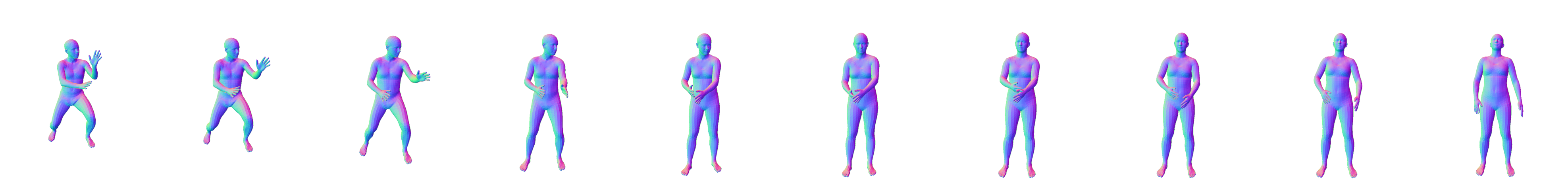}}\\
    {\includegraphics[width=0.8\textwidth]{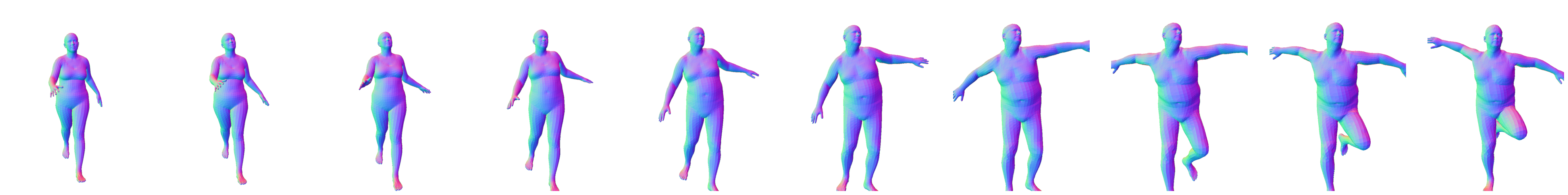}}\\
    {\includegraphics[width=0.8\textwidth]{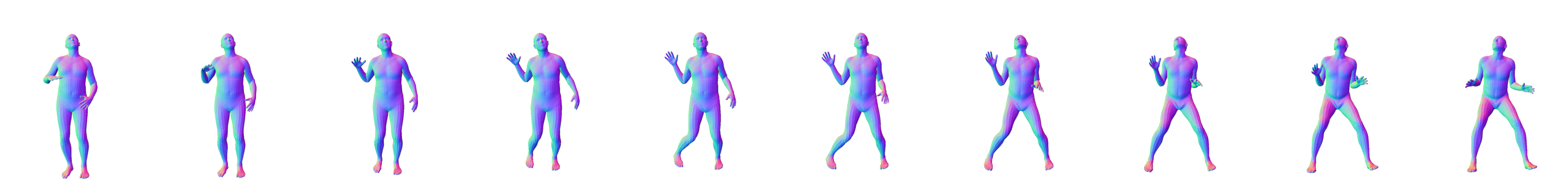}}

    \end{center}
    \vskip-6pt
        \caption{Examples of interpolating between random points in the latent space of GAN-$\beta\Theta$ mixed model. Pose and shape are entangled with each other, which complexifies the usage of such a mixed model in applications.
    }
    \label{fig:poseshape}
\end{figure*}


\begin{figure*}[t]
    \begin{center}
    {\includegraphics[width=\textwidth]{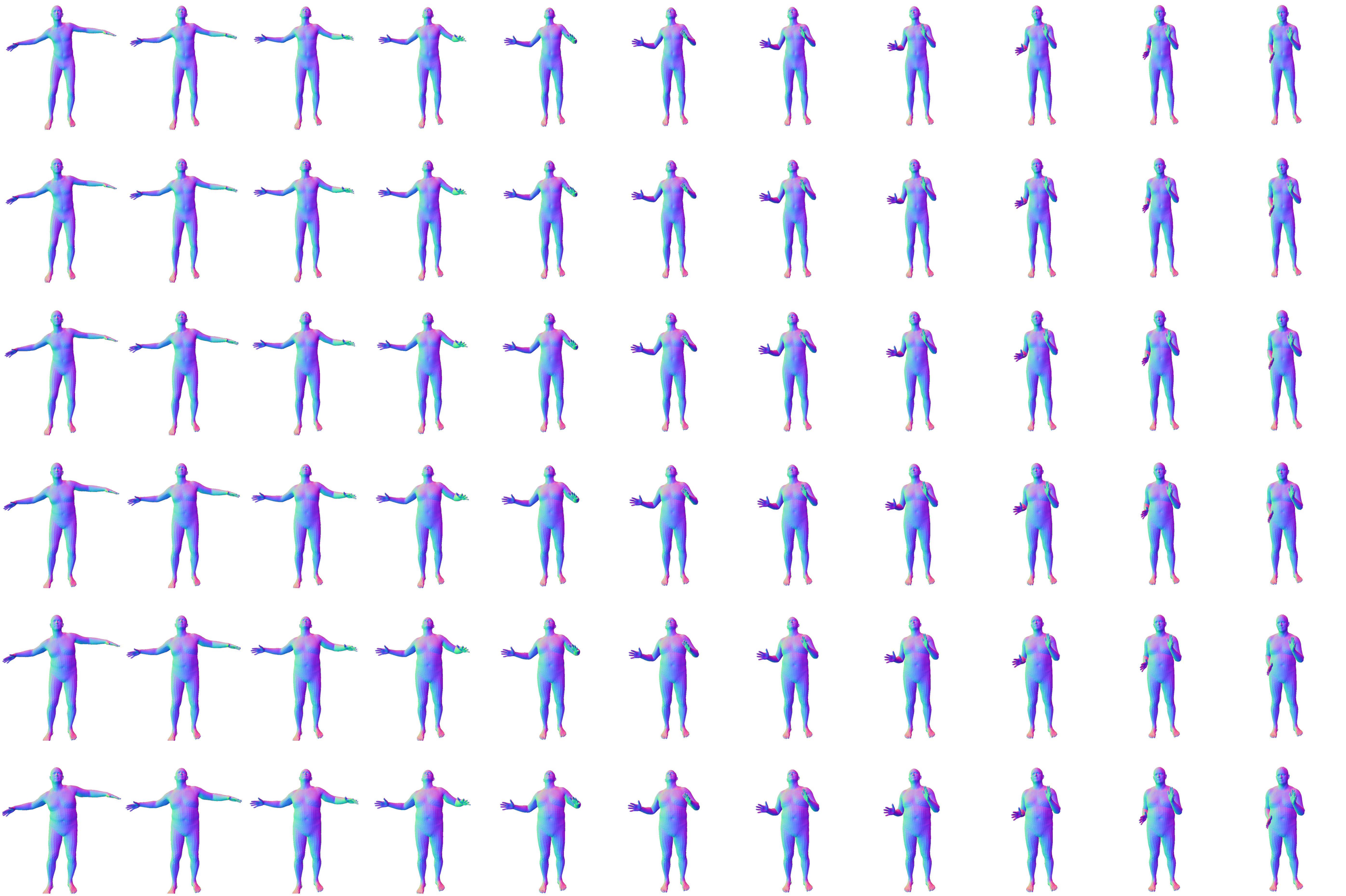}}

    \end{center}
    \vskip-6pt
        \caption{Example of generating bodies via two independent models, GAN-$\Theta$ and GAN-$\beta$. \textit{Pose} varies along axis $x$, \textit{Shape} varies along axis $y$.
        Having indepedent latent models allows to disentangle shape and pose for generating body. However, as pose and shape are in fact dependent, it might lead to generating implausible bodies (see \textit{last column, bottom rows}).
    }
    \label{fig:poseshape_disentagled}
\end{figure*}


\begin{figure*}%
    \centering
    \subfloat[\centering t-SNE on GAN-$\Theta$ (pose only, main paper).]{{\includegraphics[width=0.22\textwidth]
    {figs/tsne_ganS2.png} }}%
    \subfloat[\centering t-SNE on GAN-$\beta$ (shape only).]{{\includegraphics[width=0.22\textwidth]
    {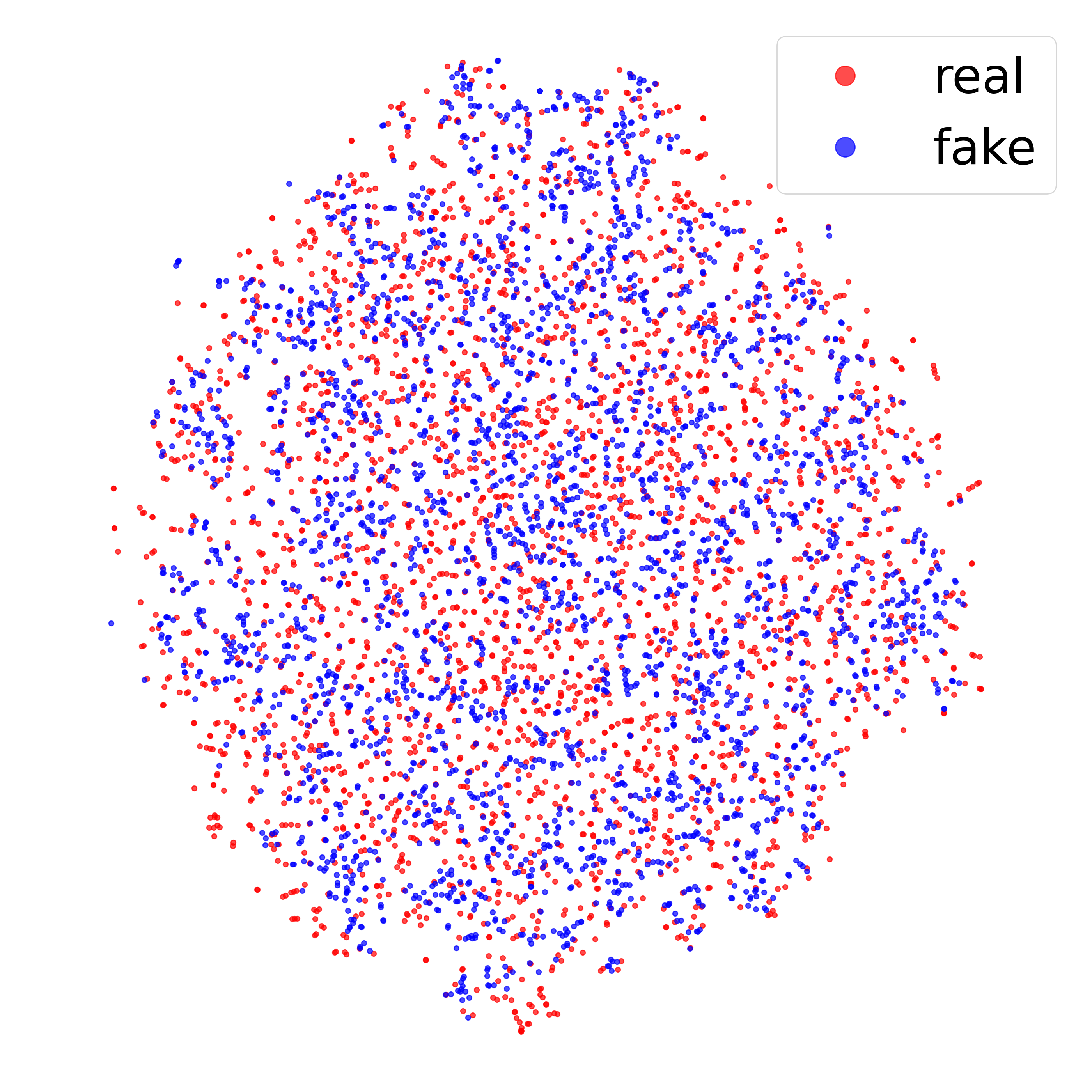} }}%
    \subfloat[\centering t-SNE on GAN-$\beta\Theta$ (pose and shape). The cluster of outliers (\textit{black rectangle}) corresponds to implausible bodies.]{{\includegraphics[width=0.22\textwidth]
    {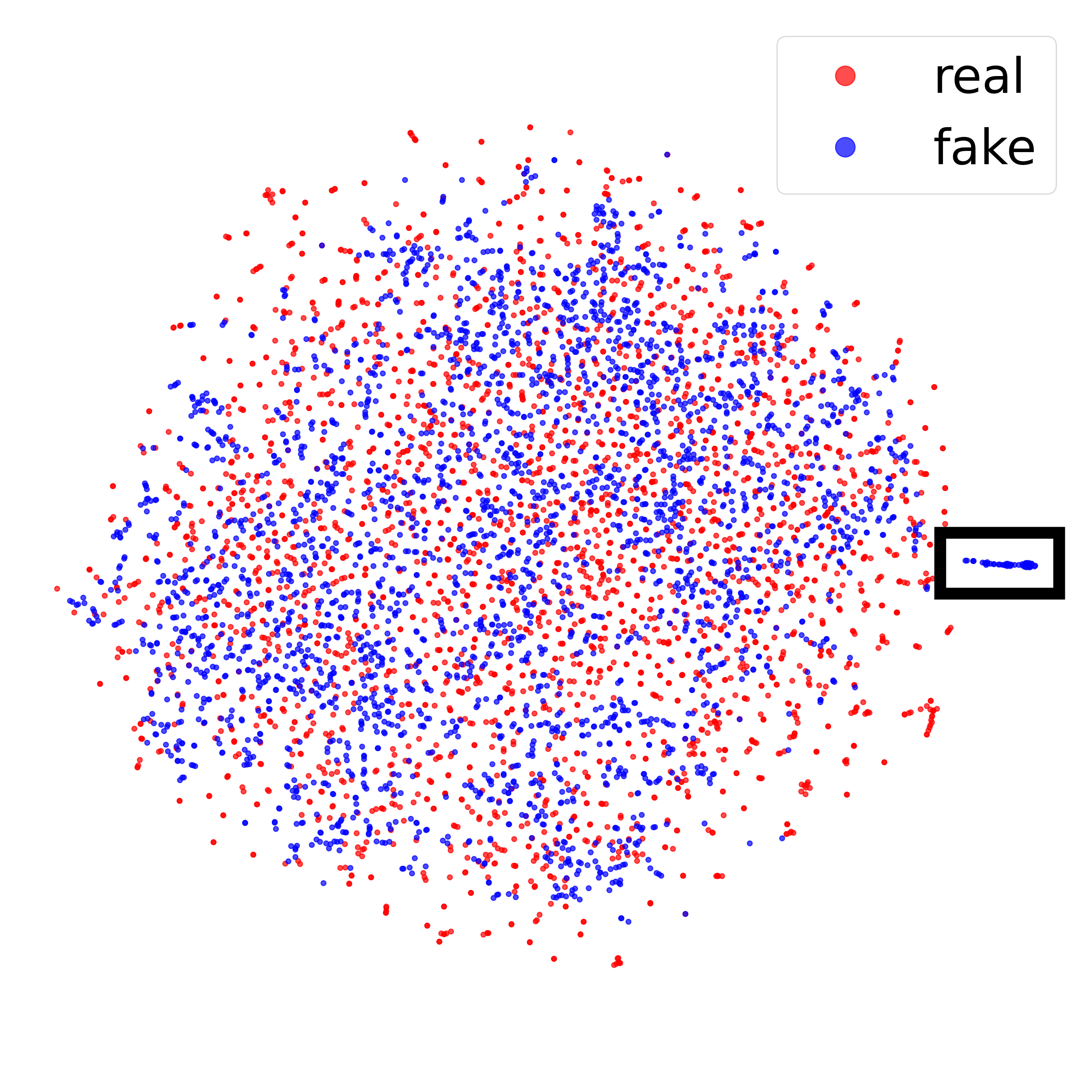} }}%
    \subfloat[\centering Unrealistic samples of GAN-$\beta\Theta$ (from the \textit{black rectangle} in (c) ). ]
    {{\centering\includegraphics[width=0.3\textwidth]
    {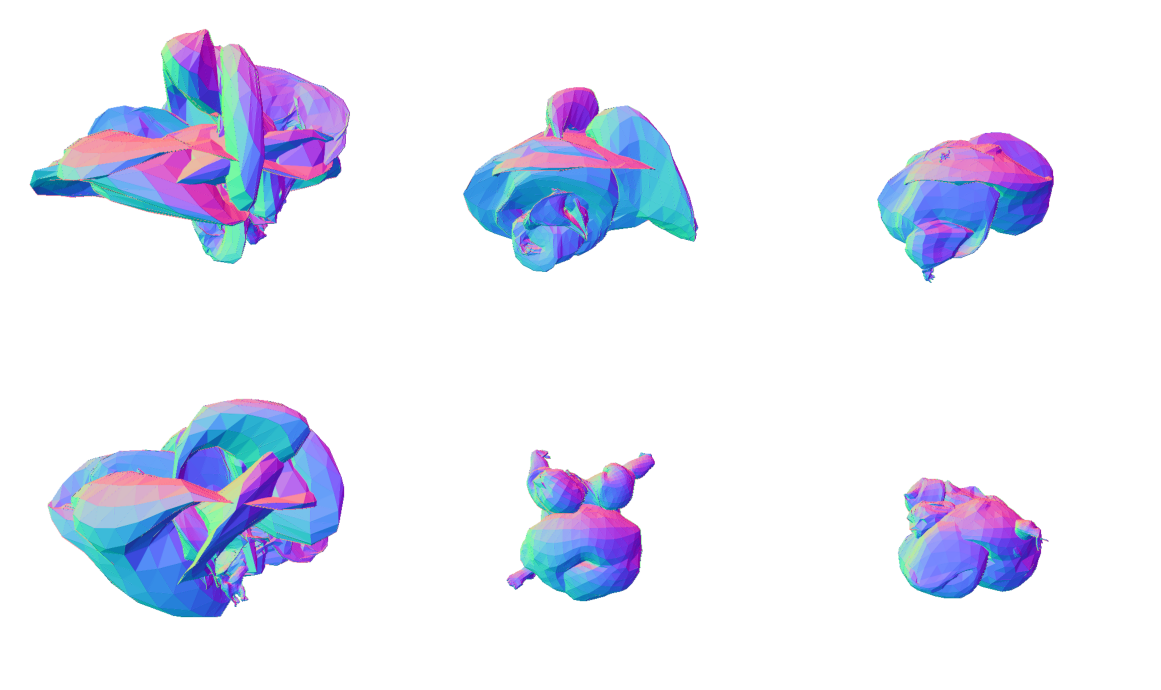} }}\label{subfig:blobs}%
    \caption{t-SNE of samples from the entangled GAN-$\beta\Theta$ (c) identifies a cluster of generated points that are not consistent with real samples. This cluster consists of outliers, shown in (d), that correspond to unrealistic bodies. t-SNE plots for independent models, GAN-$\Theta$ (a) and GAN-$\beta$ (b), demonstrate consistent data coverage. 
    }%
    \label{fig:tsne_blobs}%
\end{figure*}

The SMPL model~\cite{Loper15} is parametrized by two sets of parameters: pose $\Theta$ and shape $\beta$.
In the main paper, we explore the properties of the former for a fair comparison with VPoser~\cite{Pavlakos19} that only explores the pose prior, while the latter stays out of context.
Despite having relatively constrained shape parameters in SMPL, by using a data-driven PCA model, the unbounded nature of PCA does not restrict the SMPL output to always be plausible.

Luckily, the GAN-based approach that we used for poses can easily be applied for shapes as well. The only difference is the choice of architecture for the shape discriminator.

To provide a joint model for pose and shape, we explore two GAN variants (using spherical input space prior):

\begin{itemize}
    \item Disentangled ``shape-only" (``GAN-$\beta$") and ``pose-only" (``GAN-$\Theta$") models.
    \item Entangled ``shape+pose" ($\bz \mapsto \beta,\Theta$), denoted as ``GAN-$\beta\Theta$".
\end{itemize}

In the first case, we train an independent shape model GAN-$\beta$ and use it together with GAN-$\Theta$, while in the second case, we train both jointly, using a shared $\bz$ space.
Note that in the main paper all GAN-based models are of a kind ``GAN-$\Theta$", as they map input points to the SMPL-pose space. 

It is important to note that for training GAN-\{S,U,N\} models we used AMASS~\cite{Mahmood19} dataset, which contains an abundant number of body poses. However, the number of subjects (different body shapes) is very few ($346$ in total).  To overcome this issue, for training shape-oriented models we choose another dataset, SURREAL~\cite{Varol17}, which is composed of synthetic SMPL bodies.
In SURREAL, the shapes $\beta$ are sampled from CAESAR dataset~\cite{Pishchulin17}, which contains about $4k$ different shapes. This number is still very limited, compared to tens of millions of $\Theta$-poses in AMASS, however, we are still able to train generative models as a proof of concept.

\subsection{Disentangled GAN-$\beta$ and GAN-$\Theta$} \label{subsec:gan_shape}

Trained GAN-$\beta$ ($\bz_1 \mapsto \beta$) coupled with trained GAN-$\Theta$ model ($\bz_2 \mapsto \Theta$, from the main paper) might serve as a full prior of the original SMPL~\cite{Loper15} model. This approach learns disentangled priors for pose and shape, in the same spirit as SMPL that represents pose and shape parameters independently. However, the realistic pose and shape are \textit{not} completely disentangled. It means that pose and shape independently may account for plausible humans but combined together give a body with self-interpenetrations, as illustrated in Fig.~\ref{fig:poseshape_disentagled}.

As for the architecture of the shape discriminator, we follow HMR~\cite{Kanazawa18a} and use a 2-layer MLP.

\subsection{Entangled GAN-$\beta\Theta$}

To train a GAN for pose and shape together, we use the discriminators of GAN-$\Theta$ (GAN-S in the main paper) and GAN-$\beta$ together, and train a generator with a shared input space $\bz$. In this model, we also use a ``$\beta+\Theta$" discriminator that penalizes the full generated SMPL vector. In total we have $K+1(\text{pose}) + 1(\text{shape}) + 1(\text{pose+shape})$ discriminators.

The examples of random interpolations in the input space of $\bz$ can be found in Fig.~\ref{fig:poseshape}. As pose and shape are entangled together, it prevents utilizing such a mixed model in applications where one of these characteristics needs to remain fixed.

At the same time, not every sample corresponds to a plausible body (even with self-interpenetrations allowed). Our experiments show that generated samples might not look as humans at all. We demonstrate it in Fig.~\ref{fig:tsne_blobs} with corresponding t-SNE visualizations for each GAN model.

\section{Future work}

In most situations, independently sampling pose and shape parameters will result in realistic bodies. However, this is not always the case.
Consequently, the task of generating plausible bodies when mixing shape and pose together needs to be further investigated. It might be resolved, for example, by using the formulation of a conditional GAN, where generating pose depends on some shape features or vice versa. We explore these aspects in our further work.

\end{document}